\documentclass[10pt,journal,compsoc]{IEEEtran}
\usepackage{graphicx}
\usepackage{amsmath,amssymb} 
\usepackage{color}
\usepackage{array}
\usepackage{colortbl}
\usepackage{algorithm2e}
\usepackage{subfig}
\usepackage{epstopdf}
\usepackage{xcolor}
\ifCLASSOPTIONcompsoc
  \usepackage[nocompress]{cite}
\else
  \usepackage{cite}
\fi
\ifCLASSINFOpdf
\else
\fi
\begin{document}
%
\title{Clickstream analysis for crowd-based object segmentation with confidence}
%
%
%
%

\author{Eric~Heim,
        Alexander~Seitel,
        Jonas~Andrulis,
        Fabian Isensee,
        Christian Stock,
        Tobias Ross
        and~Lena~Maier-Hein
\IEEEcompsocitemizethanks{\IEEEcompsocthanksitem E. Heim, A. Seitel, T. Ross and L. Maier-Hein work at the Division of Computer Assisted Medical Interventions, German Cancer Research Center (DKFZ),
Heidelberg, Germany.\protect\\
E-mail:\{e.heim, a.seitel, t.ross, l.maier-hein\}@dkfz.de
\IEEEcompsocthanksitem F. Isensee works at the Division of Medical Image Computing, German Cancer Research Center (DKFZ),
Heidelberg, Germany.\protect\\
E-mail:f.isensee@dkfz.de
\IEEEcompsocthanksitem J. Andrulis works at Pallas Ludens GmbH,
Heidelberg, Germany.\protect\\
E-mail:jonas.andrulis@pallas-ludens.awsapps.com
\IEEEcompsocthanksitem C. Stock works at the Divison of Clinical Epidemiology and Aging Research at the German Cancer Research Center (DKFZ), Heidelberg, Germany, and the Institute of Medical Biometry and Informatics, University Hospital Heidelberg, Heidelberg, Germany.\protect\\
E-mail:stock@imbi.uni-heidelberg.de
}

}

%
%

\markboth{to appear in IEEE Transactions On Pattern Analysis and Machine Intelligence, DOI 10.1109/TPAMI.2017.2777967}%
{Shell \MakeLowercase{\textit{et al.}}: Bare Demo of IEEEtran.cls for Computer Society Journals}
%

\IEEEpubid{0162-8828~\copyright~2017 IEEE. See http://www.ieee.org/publications\_standards/publications/rights/index.html for more information.}


\IEEEtitleabstractindextext{%
\begin{abstract}
With the rapidly increasing interest in machine learning based solutions for automatic image annotation, the availability of reference annotations for algorithm training is one of the major bottlenecks in the field. Crowdsourcing has evolved as a valuable option for low-cost and large-scale data annotation; however, quality control remains a major issue which needs to be addressed. To our knowledge, we are the first to analyze the \emph{annotation process} to improve crowd-sourced image segmentation. Our method involves training a regressor to estimate the quality of a segmentation from the annotator's clickstream data. The quality estimation can be used to identify spam and weight individual annotations by their (estimated) quality when merging multiple segmentations of one image. Using a total of 29,000 crowd annotations performed on publicly available data of different object classes, we show that (1) our method is highly accurate in estimating the segmentation quality based on clickstream data, (2) outperforms state-of-the-art methods for merging multiple annotations. As the regressor does not need to be trained on the object class that it is applied to it can be regarded as a low-cost option for quality control and confidence analysis in the context of crowd-based image annotation.
\end{abstract}

}

\maketitle
\IEEEpeerreviewmaketitle

\IEEEdisplaynontitleabstractindextext

%
\IEEEpeerreviewmaketitle

\IEEEraisesectionheading{\section{Introduction}\label{sec:introduction}}

\IEEEPARstart{T}{he} interest in machine learning techniques for data processing has been rapidly growing in various fields including the automotive industry \cite{Luckow2015}, computer vision \cite{imagenet_cvpr09}, and biomedical image processing \cite{Zhou2015}. The major bottleneck of most of those techniques - especially with the rise of deep learning algorithms - is the annotation of the often large amount of required training data \cite{krizhevsky2012imagenet}. Crowdsourcing has become popular in this context as it is based on outsourcing cognitive tasks to many anonymous, untrained individual \emph{workers} from an online community \cite{albarqouni2016aggnet,raykar2010learning}. It thus provides a valuable tool for low-cost and large-scale data annotation. One of the major challenges in crowd-based image annotation is quality control. Although many workers are highly skilled and motivated, the presence of \textit{spammers} is a severe problem as they are mainly interested in receiving the reward for a given task by investing the minimum amount of time \cite{kazai2011worker}. As detailed in section \ref{sec:soa}, methods proposed to address this issue have led to better overall results. However, they either require expert workers to perform time-wasting tasks on images for which the reference is already known, use inter worker agreement on segmentations of multiple workers on the same image or restrict the pool of potential workers to those that have a history of exceptionally good rating by the task providers (and thus reduce annotation speed). These quality control approaches all rely on the annotation result. 

\begin{figure}[!!!htb]
  \centering
  \includegraphics[width=0.99\linewidth]{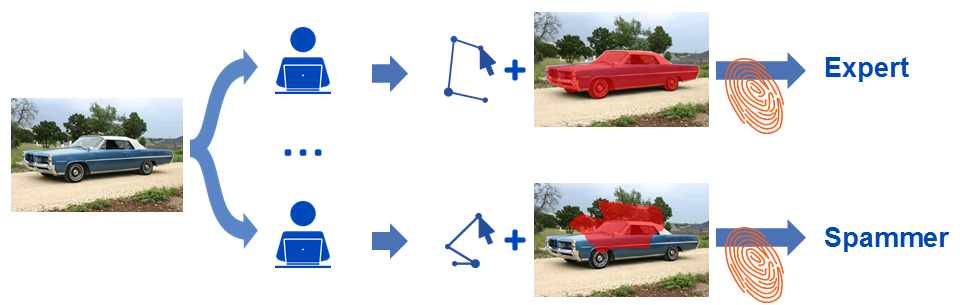}
  \caption{Annotation process analysis based segmentation quality estimation in crowd-sourced image segmentation. The quality of the segmentation is derived from the worker's mouse actions recorded in the clickstream. }
    \label{fig:concept_spam_detection}
\end{figure}

The hypothesis of this paper is that the quality of a segmentation is reflected in the way a worker annotated the image (Fig. \ref{fig:concept_spam_detection}). Specifically, we assume that spammers behave differently (e.g. have a faster mouse movement) compared to workers that produce high-quality annotations. Our work therefore focuses on the annotation process itself to estimate the quality of crowd-sourced image segmentations. For this purpose, all user traces during the annotation process are recorded as \textit{clickstreams}. A clickstream consists of consecutive time stamped events triggered by a user while interacting with a web application. 

To our knowledge, we are the first to use \textit{annotation process-based features} for quality control in crowd-sourced image segmentation. Our contribution is illustrated in Fig. \ref{fig:training_overview}-\ref{fig:testing_overview} and comprises (1) an approach to estimate the segmentation quality based on clickstream analysis and (2) a method for accurate crowd-based object segmentation based on confidence-based weighting of individual crowd segmentations.
The outline of the paper is as follows: Section \ref{sec:soa} provides an overview of related work that has previously been published on this topic. The quality estimation concept for crowd-sourced image segmentation is detailed in section \ref{sec:methods}, including the description of the performed validation.
Finally, experimental results are presented and discussed in sections \ref{sec:results} and \ref{sec:discussion} respectively.

 

\section{Related work} \label{sec:soa}
We review related work in view of the two main research fields related to our approach. Section \ref{sec:soa_crowdsourcing} discusses the methods proposed for quality control in crowdsourcing, section \ref{sec:soa_clickstream} summarizes the research on user behavior analysis outside the field of crowdsourcing.

\subsection{Quality control in crowdsourcing tasks}\label{sec:soa_crowdsourcing}
Approaches to quality control in crowd-sourced data annotation can roughly be divided into the following categories:

\subsubsection{Integration of reference data into annotation task}
Von Ahn's reCaptcha \cite{von2008recaptcha}, one of the first approaches to ensure the quality of crowd-sourced annotations, proposes pairing an unknown label with a label for which a reference is available (captcha) and rejecting the new label when the performed captcha has failed. For the final result, the consensus of labels from different workers is used. Even though this is an online approach, where false labels are directly rejected, it requires pre-labeled reference data and is not cost-effective, since the crowd has to additionally annotate already known labels, which is time consuming, especially in the context of object segmentation. In an initial quality control step, Lin et al. \cite{lin2014microsoft} rejected workers based on their performance in segmenting images of a training task and verified trusted worker's annotations via a manual verification step. This approach still relies on reference data and does not incorporate the expertise of specific workers. Kazai et al. \cite{kazai2011worker} conducted worker behavioral analyses and grouped them into spammers, sloppy, incompetent, competent and diligent workers by a worker survey and crowd annotations of already labelled digitized books. They did not explore the application of the worker's expertise on other task types such as image annotation. An approach to categorize the workers into spammers, workers who did not understand the task, and reliable workers from their annotation on reference images was presented by Oleson et al. \cite{Oleson2011}. Cabezas et al. \cite{Cabezas2015} proposed a quality control approach using interactive object segmentation based on background and foreground clicks in the image. Incorrect clicks are detected by their spatial neighborhood inside superpixels. Finally the clicks were weighted based on the worker's expertise. The workers had to perform segmentations on gold standard images to determine their level of expertise. However the approach only uses an initial expertise estimation on a reference task and can thus not assure the quality of each individual label during the annotation process.

\subsubsection{Majority voting}
\label{sec:majorityVoting_SOA}
Another approach to detect low-quality labels is to acquire multiple labels from each sample and rank them against the majority of the acquired labels \cite{imagenet_cvpr09,activeLearningObjectDetection,von2004labeling,maierh14Masses,gurari2015collect}. A pixel of the image to be annotated is classified as belonging to the object if the majority of workers have classified it as object (e.g. \cite{maierh14Masses}). The assumption behind the majority voting approach is that the majority of labels are of good quality, which is likely to result in a larger amount of labelling data being acquired from the crowd. When merging a majority of accurate object segmentations, majority voting can provide solutions close to the optima\cite{wang2009classifier}.  

\subsubsection{Manual grading of annotation quality}
\label{sec:manualgraqding_SOA}
A commonly used means for ensuring quality of crowd-sourced annotations is to let peer crowd-workers rate the quality of performed annotations. This is among others integrated in the annotation pipelines for generating large image databases such as LabelMe \cite{Russell2008} and COCO \cite{lin2014microsoft}. This manual verification task imposes additional annotation costs that could be avoided when quality estimation is performed automatically.

\subsubsection{Automatic annotation quality estimation}
Vittayakorn et al. \cite{Vittayakorn2011} investigated several scoring functions to assess the quality of crowd-sourced object annotations. Quality measures include features derived from the image itself (e.g. image edges) and parameters obtained from the resulting annotation contour or segmentation (e.g. number of control points, annotation size). Features related to the actual \textit{process} of performing the annotation have not been applied for quality control. Hence, the method does not rely on recording the user’s clickstream. Welinder et al. \cite{Welinder2010} proposed a general method to determine a reference value of some properties from multiple noisy annotations; this is a method which inherently evaluates the annotator's expertise and reliability. An expectation maximization (EM) approach is used to infer the target annotation as well as a confidence value for the worker's reliability. This method has currently only been tested on simple binary decisions and bounding boxes. Similarly, Long et al. \cite{Long2016} introduced a Joint Gaussian Process Model for active visual recognition and expertise estimation. This approach has also not been tested on more complex annotations such as object segmentations. A Bayesian probabilistic model for rating the competence of workers creating binary labels of images out of multiple label aggregations per image from different workers was presented by Welinder et al. \cite{welinder2010multidimensional}. Several methods have been proposed that estimate the annotation quality by training a regressor on features derived from the image and/or final annotation contour \cite{Gurari2016,Carreira2010,Arbelaez2014,Jain2013,Jain2016}. Sameki et al. \cite{Sameki2015} recently presented a method that detects the worker's behavior from the number of mouse clicks, the time per task and the time per user click. They were able to show that the number of contour points in the final annotation and the annotation time are good predictors for segmentation quality. The authors, however, state that the improvements to be gained from this limited number of features might be minimal for certain applications such as the segmentation of bio-medical images. 

Other quality control approaches in crowd-based image annotation have been designed for specific applications, such as correspondence search in endoscopic images \cite{maierh2014} and cannot be generalized to other tasks, such as segmentation. Mao et al. \cite{mao2013volunteering} investigated how different payment methods are related to the task outcome and came to the conclusion that financial incentives can be used to boost annotation speed at the cost of quality.

\subsection{User behavior analysis}\label{sec:soa_clickstream}
Several groups have investigated using clickstream data or mouse dynamics for user behavior analysis outside the field of crowdsourcing. Successful approaches have been presented to identify specific users by making use of biometric features based on mouse dynamics \cite{ahmed2007new,Feher2012} for user authentication and intrusion detection in computer systems. In Ahmed et al. \cite{ahmed2007new} a neural network is trained to recognize a specific user via a feature set of mouse dynamics, e.g. velocity, acceleration, clicks, moving direction. The method of Feher et al. \cite{Feher2012} extended this approach by using a random forest classifier trained on a similar features and combining the results of individual mouse actions in contrast to the histogram-based approach by Ahmed et al.\cite{ahmed2007new}. While these approaches have been widely used for user authentication and intrusion detection, their application to segmentation and annotation tasks has not yet been presented. A method to detect fake identities and sybil accounts in online communities, based on features extracted out of clickstreams, was presented by Wang et al. \cite{sybils}. In contrast to the mouse dynamics based approaches, user actions such as uploading photos, chatting with friends, pressing a specific button, etc. are used here to train the classifier. Wang et al. \cite{wang2016unsupervised} showed that clickstream analysis can be used to cluster social network users based on their behaviour via an unsupervised learning approach. Lee et al. \cite{lee2015characterizing} presented a machine learning based approach to detect content in social networks that was manipulated through crowdsourcing. They trained a classifier on social network features, e.g. number of friends, posts on websites, overall activity time in the social network and achieved a high level of accuracy in detecting crowd manipulated content. Rzeszotarski et al. \cite{rzeszotarski2011instrumenting} presented a user-behavior-based method for estimating the quality of crowdsourcing for classification tasks, comprehensive reading tasks and text generation tasks. 

\bigskip
In conclusion, while there has been a considerable effort in controlling the quality of crowd-sourced annotations and analysing user behaviour, we are not aware of any prior work on using annotation process-based features for quality control in crowdsourcing. Furthermore, confidence-based annotation merging using quality estimation has not been introduced to date.

\section{Methods} \label{sec:methods}
This section presents the concept of our confidence-based method for annotation merging (Sec. \ref{sec:concept}), a prototype implementation of the concept (Sec. \ref{sec:implementation}) and the experimental design of our validation study (Sec. \ref{sec:validation}).

\subsection{Segmentation concept} \label{sec:concept}
\begin{figure}[!!!htb]
  \centering
  \includegraphics[width=\columnwidth]{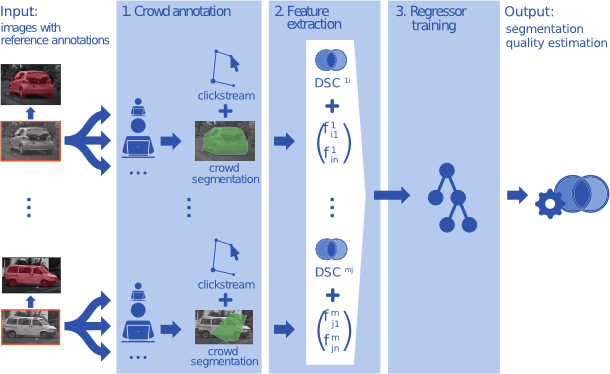}
  \caption{Training of segmentation quality estimator. Initially, images with known reference segmentations are distributed to multiple crowd workers. While the workers are segmenting the images, the system records their annotation behavior (clickstreams). For each annotated image, the clickstream is converted into a feature vector characterizing the worker's interaction behavior. The DSC is computed using the reference annotation. The set of all collected feature vectors with corresponding DSC values is then used to train a regressor to estimate the DSC solely based on a worker's clickstream. }
    \label{fig:training_overview}
\end{figure}

\begin{figure}[!!!htb]
  \centering
  \includegraphics[width=\columnwidth]{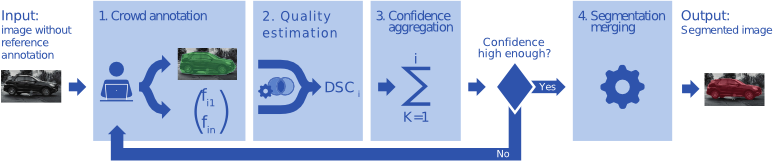}
  \caption{Concept for crowd-based image segmentation based on a trained segmentation quality estimator (Fig. \ref{fig:training_overview}). The image to be annotated is repeatedly distributed to the crowd until a certain confidence level is reached. The obtained segmentations are merged in a weighted manner, where the weight of the worker's annotation increases with the estimated DSC on that specific image.}
    \label{fig:testing_overview}
\end{figure}

The purpose of this contribution was to develop a quality control method for crowd-based object segmentation that does not rely on (1)  additional tasks (with known outcome) to be performed or (2) prior knowledge of a specific worker's annotation history. Inspired by previous work on clickstream analysis for user identification \cite{sybils,wang2016unsupervised}, the hypothesis of our work is that the clickstream data itself is sufficient to estimate a worker's performance. Our concept involves a training process (Fig. \ref{fig:training_overview}), in which a regressor is trained to estimate the quality of a given segmentation by using features extracted from clickstream data. To segment an unseen image, the image is repeatedly distributed to the crowd until a certain number of segmentations with a high estimated quality is reached (Fig. \ref{fig:testing_overview}). The DICE similarity coefficient (DSC) \cite{dice1945measures} is used as a measure for segmentation quality, which is defined by comparing a segmentation $U$ to its corresponding reference segmentation $V$ (Eq. \ref{eq:dice}).
\begin{equation} \label{eq:dice}
 DSC=\frac{2|V \cap U|}{|V|+|U|}
\end{equation}
The obtained segmentations are then merged in a weighted manner, according to their estimated quality. An implementation of this concept is presented in the following section \ref{sec:implementation}.
 
\subsection{Prototype implementation} \label{sec:implementation}
With our prototype implementation, we focus on single-object segmentation. The prototype comprises a user interface for crowd-based image segmentation (Sec. \ref{sec:ui}) as well as capabilities for clickstream data collection (Sec. \ref{sec:clickstreamdata}), feature extraction (Sec. \ref{sec:featureExtraction}), segmentation quality estimation (Sec. \ref{sec:prediction}) and confidence-based annotation merging (Sec. \ref{sec:confidence_voting}).

\subsubsection{User Interface} \label{sec:ui}
The prototype implementation incorporates a web-based user interface implemented in HTML and JavaScript to perform the image segmentations and collect the corresponding clickstream data. It provides basic functionalities to create, delete and correct contours. The worker can draw a contour by either pressing and holding the left mouse button while dragging the cursor or define points via clicks on the canvas. These points are then successively connected by lines, resulting in the segmentation contour. Typically, workers use a combination of both modes, e.g. if the object contains curvy regions combined with sharp corners or long lines, the worker might only set points to draw lines or corners and continues drawing the curves by dragging the mouse and vice versa. To help create accurate contours it is possible to zoom into the image by using the mouse wheel. If the worker is not satisfied with the created segmentation, the contour can be corrected by selecting and dragging single points to the desired position or by deleting redundant points by a double click. It is also possible to delete the complete contour and restart the segmentation from scratch.

\subsubsection{Clickstream data collection} \label{sec:clickstreamdata}
During the segmentation task, every action triggered by the worker's mouse is saved to the clickstream. In addition to worker-triggered events, the current mouse position is continuously recorded. We represent the clickstream as a sequence of successively occurring mouse events $E = \lbrace e_1, \cdots , e_i \rbrace$. The clickstream events are sorted by their time stamp $t_i$ in their recorded chronological order. Each recorded event provides $x$ and $y$ coordinates of a point $\vec{p} \in \mathbb{R}^2$ in the canvas coordinate system, their corresponding coordinates transformed into the image coordinate system, the event type and the object upon the event was triggered. We differentiate between the following event types and actions: \textit{mouse-down}, \textit{mouse-up}, \textit{mouse wheel}, \textit{double-click}, \textit{mouse-move}.
The following objects are present in the user interface: \textit{canvas}, \textit{delete contour button}, \textit{zoom button}, \textit{save button}.

\subsubsection{Feature extraction} \label{sec:featureExtraction}
Based on the assumption that reliable workers will interact differently with the program than malicious workers, we derive our feature set from the recorded clickstream. We assume that reliable workers will put more time and effort into creating accurate segmentations than malicious workers will do. Furthermore, we assume that the worker's behavior will change based on their level of expertise. Trained experts, for example, might create high-quality segmentations in less time with less effort than inexperienced, untrained workers. Inexperienced workers might instead create a high amount of user input that will not necessarily result in accurate segmentations.
In order to estimate the segmentation quality, we need a feature set that is able to classify the quality of segmentations of different object types without incorporating knowledge of the underlying system, object types or worker identity.
We define the following feature set that includes features calculated from both the clickstream and the image itself.

\subsubsection*{Clickstream-based features} \label{sec:clickstreamFeatures}

\paragraph*{\textbf{Velocity}} $\forall \ e_i \in E$ we compute a velocity vector $\vec{v}_i$ based on the elapsed time $\Delta t_i$ (in milliseconds) and travelled distance $\Delta \vec{p}_i$ on the canvas with the positions $\vec{p}_{i}$ and $\vec{p}_{i-1}$ of two successive events $e_i$ and $e_{i-1}  \in E$:
\begin{equation} \label{eq:velocity}
\vec{v}_i = \frac{\Delta \vec{p}_i}{\Delta t_i} = \frac{\vec{p}_{i-1} -\vec{p}_i}{t_{i-1} -  t_i}
\end{equation}
The velocity is only computed for mouse-move operations. On mouse or button clicks the velocity is set to zero. The mean, median, standard deviation and 95\% quantile of the velocity $\forall \ e_i \in E$ are used as features.

\paragraph*{\textbf{Acceleration}} The acceleration $\vec{a}_i$ of every event in the clickstream is derived from the velocity change $\Delta \vec{v}_i$ and elapsed time $\Delta t_i$ 
between two successive events $e_i \in E$ and $e_{i-1} \in E$:
\begin{equation} \label{eq:acceleration}
 \vec{a}_i =  \frac{\Delta \vec{v_i}}{\Delta t_i}
\end{equation}
The mean, median, standard deviation and 95\% quantile of the acceleration $\forall \ e_i \in E$ are used as features.
\paragraph*{\textbf{Zoom}}
We extract the total number of zoom events out of the clickstream by using the event and object types detailed in Sec. \ref{sec:clickstreamdata}.
\paragraph*{\textbf{Canvas clicks}}
The total number of mouse-clicks that are executed on the canvas are used as a feature.
\paragraph*{\textbf{Double clicks}}
The total number of double-clicks that are executed on the canvas are used as a feature.
\paragraph*{\textbf{Elapsed time}}
The duration of the whole task is calculated as the difference between the time stamps of the first and last event in the clickstream:  $\Delta T = t_n - t_0$. The time is normalized by the mouse clicks and actions as described in \cite{Sameki2015}, i.e.: $\frac{\Delta T}{canvas\ clicks}$. 
\paragraph*{\textbf{Ratio of traveled mouse distance and length of the segmented contour}} In contrast to the absolute size of a contour used by Vittayakorn et al. \cite{Vittayakorn2011}, we assume that the length of a created contour will be in relation to the total traveled mouse distance in the image space. We assume that a spammer will typically try to create a random contour on the canvas that will be similar in length to the total traveled distance, whereas reliable users will more likely perform different mouse movements such as zooming or moving the mouse to adjust the view or correcting created contours.
\paragraph*{\textbf{Mouse strokes}} The total number of mouse strokes is used as a feature. We define a mouse stroke as a sequence of mouse-move events $S \subseteq E$ that occur between a mouse 
down and up event. As explained in section \ref{sec:ui}, mouse strokes are used to draw a contour or to select and drag points to correct existing contours. 
\paragraph*{\textbf{Draw operations and contour correction}}
To distinguish whether a mouse stroke is used to draw a new contour or correct an existing one, the clickstream is processed using Algorithm \ref{alg:stream_process} based on the event types and objects previously described.

\begin{algorithm} 
 \KwData{mouse strokes $S$, clickstream $E$}
 \KwResult{set of draw operations $D$, set of corrections $K$ }
 \Begin{
 $i \longleftarrow 0$\;
 \For{$S_i \subseteq E$}{
   \eIf{$S_{i}.down.\vec{p} \equiv S_{i-1}.up.\vec{p},\ with \ S_{i-1} \in D$} {
      $D.add(S_i)$\;
   }{
      \eIf{$\exists \ e \in E :S_i.down.\vec{p} \equiv e.\vec{p} \land S_{i}.t \leq e.t$}{
         $K.add(S_i)$\;
      }{
        $D.add(S_i)$\;
      }
    }
   }
  }
\caption{Clickstream processing to identify if a mouse stroke $S$ is a draw operation $D$ or a correction $K$. We distinguish between three different cases: (1) If the mouse down event of the current mouse stroke occurs at the same position as the mouse up event that terminated the last draw event, the current mouse stroke $S_i$ is considered as a drawing event. (2) If this did not occur and the down event of the current stroke occurred at the same position as a previous event in the clickstream, the current mouse stroke $S_i$ is considered as a correction event. (3) Otherwise a new contour is started and the current mouse stroke $S_i$ is a draw event. A kd-tree \cite{KDtree} is used to speed up the search of events in the clickstream based on their canvas position.}
\label{alg:stream_process}
\end{algorithm}
We use the absolute number of draw events and executed corrections as features. In addition we use the acceleration and velocity for the executed draw and corrections and use the mean, median, standard deviation and the 95\% quantile as features.

\subsubsection*{Image-based features} \label{sec:image_features}
Similar to Vittayakorn et al. \cite{Vittayakorn2011}, we assume that for an accurate segmentation, the contour will mainly be located on or next to image edges. While creating a contour, the worker will try to follow edges in the image and the mouse will move perpendicular to the gradient direction. The contour of a spam\-mer, who is not segmenting an object, will, in contrast, not be created perpendicular to the gradient direction. In addition to the relation between the mouse-move direction and the gradient, the quality of the resulting contour is ranked based on the gradients and the interpolated vertex normals of the created polygon. In this case, we assume that vertex normals are collinear to the gradient direction.

We thereby define features that are based on the image gradient and compute the angle $\gamma_i$ between the gradient direction $\vec{g}_{x,y}$ at the image coordinates $x,y$ and the mouse-move direction $\vec{d_i}$ for each event $e_i$. The gradient is computed using recursive Gaussian filtering \cite{recursivegaussian} as implemented in the Insight Toolkit (ITK) \cite{ITKSoftwareGuideSecondEdition} while the mouse-move direction $\vec{d}_i$ is derived by normalizing the velocity $v_i$ for the event $e_i$: 
\begin{equation} \label{eq:movedirection_norm}
\vec{d}_i = \frac{\vec{v_i}}{\|\vec{v}_i \|} 
\end{equation}
In the first step we compute the angle between $\vec{g}_{x,y}$ and $\vec{d_i}$ according to Eq. \ref{eq:angle}: 

\begin{equation} \label{eq:angle}
 \omega(\vec{d_i},\vec{g}_{x,y}) = acos(\frac{\vec{d_i} \bullet \vec{g}_{x,y}}{\parallel \vec{d}_i \parallel \cdot \parallel \vec{g}_{x,y} \parallel}) \cdot \frac{180}{\pi}
\end{equation}
use the smallest angle $\gamma_i$: 
\begin{equation} \label{eq:smallest_angle}
  \gamma_i =
  \begin{cases}
    \omega(\vec{d}_i,\vec{g}_{x,y}), & \text{if}\  0^\circ \leq \omega(\vec{d}_i,\vec{g}_{x,y}) \leq 180^\circ \\
    360^\circ - \omega(\vec{d}_i,\vec{g}_{x,y}), & \text{otherwise}
  \end{cases}
\end{equation}
and normalize to an angle between $0^\circ$ and $90^\circ$: 

\begin{equation} \label{eq:angle_norm}
  \bar{\gamma}_i = \epsilon_{\gamma} - \lvert \epsilon_{\gamma} - \gamma_i \lvert
\end{equation}
with $\epsilon_{\gamma} = 90^\circ$. We then define the following image-based features:
\paragraph*{\textbf{Features extracted from contour drawing and correction events}}
The mean, standard deviation, median and 95\% quantile of the angles $\bar{\gamma}_i$ are calculated as features for all contour drawing events $D$, all contour corrections $K$ and all consecutive mouse click events. For the mouse click events, the direction vector $\vec{d}_i$ is calculated as the line segment connecting the current mouse click with a previous one.

\paragraph*{\textbf{Features extracted from the final contour}}
The final contour is defined as a set of consecutive connected 2D vertices represented by points in the image coordinate system $X=\lbrace{\vec{p}_1,\cdots, \vec{p}_n \rbrace}$, with $\vec{p}_i \in \mathbb{R}^2$. The normal for each line segment $\vec{n}_i$ is calculated with two consecutive vertices $\vec{p}_i,\vec{p}_{i+1} \in X$: 
\begin{equation} \label{eq:linenormal}
  \vec{n_i} =
   \begin{bmatrix}
    -1 \cdot (\vec{p}_{{i+1}_y} - \vec{p}_{{i}_y})\\
    (\vec{p}_{{i+1}_x} - \vec{p}_{{i}_x)}
   \end{bmatrix}
\end{equation}
The interpolated vertex normal $\vec{\tilde{n}}_i$ is calculated by a linear interpolation of the line segment normals of two adjacent line segments $\vec{n}_{i-1}$ and $\vec{n}_i$.
\begin{equation} \label{eq:vertexnormal}
   \vec{\tilde{n}}_i=\frac{1}{2} \vec{n}_{i-1} + \frac{1}{2} \vec{n}_{i}
\end{equation}
With the vertex normals and the gradient directions, we compute $\bar{\gamma}_i \ \forall \ p \in X$ according to Eq. \ref{eq:angle_norm} with the image gradient $\vec{g}_{x,y}$ and the interpolated vertex normal $\vec{\tilde{n}}_i$ instead of the drawing direction $\vec{d}_i$. We use the mean, standard, deviation, median and 95\% quantile of all computed angles as features.

\subsubsection{Estimation of segmentation quality} \label{sec:prediction}
With the set of features introduced in the previous Sec. \ref{sec:featureExtraction} and the DSC of segmentations for which a reference segmentation exists, we train a random forest regressor \cite{randomForest} to estimate the quality of unseen crowd segmentations. We use the random forest regressor implementation from scikit-learn \cite{scikit} to estimate the DSC $\hat{s}_j$ of an unseen crowd segmentation $U_j$. To determine the parameters for our random forest regressor we used a data set consisting of 20 images of cars that were not part of the validation data set. For each image, we obtained reference segmentations with the method described in \cite{kondermann2014stereo}. In addition, we acquired a total of 500 crowd segmentations for each image with the platform provided by the Pallas Ludens GmbH\footnote{https://pallas-ludens.com}, resulting in 10,000 segmentations for cross-validation. We determined the tree depth and the minimum number of samples per leaf of our random forest regressor by running a 10 fold labeled cross-validation optimizing the $R^2$ score with the estimated DSC $\hat{s_i}$, the corresponding real DSC $s_j$ for every segmentation and the mean DSC $\overline{s}$ of the data set (Eq. \ref{eq:r2score}):   
\begin{equation} 
\label{eq:r2score}
    R^2= 1 - \frac{ \sum^{n}_{j=1} ( s_j - \hat{s}_j )^2 }{\sum^{n}_{j=1} ( s_j - \overline{s} )^2}
\end{equation}
Before running the cross-validation we ensured that neither the same workers or the same images were included in the test and training data set simultaneously. 
For a random forest regressor with 500 trees, a minimum of three samples per leaf and extending the tree depth until all leaf nodes are pure, we achieved a mean $R^2$ score of $0.71 \pm 0.04$. 

\subsubsection{Confidence-based segmentation merging} \label{sec:confidence_voting}
To generate a new segmentation, the workers annotations were merged according to their estimated quality. We investigated two different methods for merging the segmentations: (1) A confidence-weighted majority voting approach based on the estimated segmentation quality and (2) a simultaneous truth and performance level estimation (STAPLE) algorithm based approach \cite{commowick2010incorporating}.

\subsubsection*{Confidence-weighted majority voting}
Based on the estimated quality, segmentations are discarded if the estimated DSC $\hat{s}_j$ for a segmentation $U_j$ falls under a predefined DSC threshold $\epsilon_t \in [0,1]$. 
With the estimated DSC values $\hat{s}_j$ we compute a normalized confidence value $\kappa(\hat{s}_j) \in [0,1]$ for the remaining segmentations:
\begin{equation} \label{eq:confidenceCalculation}
 \kappa(\hat{s}_j) = \frac{\hat{s}_j - \epsilon_t}{1 - \epsilon_t}
\end{equation}
We denote the two dimensional segmentations images $U_j(x,y)$ with the width $m$ and height $n$, where $(x,y)$ is the coordinate of the pixel in the image with $U_j(x,y) \in \lbrace{0,1\rbrace}$. Each segmentation $U_j(x,y)$ is weighted with the estimated confidence (Eq. \ref{eq:confidenceCalculation}) and the confidence-weighted pixel values are accumulated in the image $\Delta U(x,y)$:  
\begin{equation} \label{eq:acc_confidence}
   \Delta U(x,y) = \sum_{j=1}^{\lambda}  U_j(x,y) \cdot \kappa(\hat{s}_j) 
\end{equation}
The smallest integer value $\mu$ representing the majority of $\lambda$ segmentations is used to calculate the fraction of the maximum accumulated confidence value $\psi$ in $\Delta U(x,y)$, that is required to classify a pixel as belonging to the segmented object:
\begin{equation} \label{eq:majority_fraction}
  \psi = \frac{\max\limits_{x=1,y=1}^{m,n} \Delta U(x,y) }{\lambda} \cdot \mu
\end{equation}
The final confidence-weighted segmentation $H(x,y)$ is calculated by applying a binary decision to each pixel of the image $\Delta U(x,y)$:
\begin{equation} \label{eq:binary_confidence}
   H(x,y) = 
     \begin{cases}
     1, & \text{if} \ \Delta U(x,y) \geq \psi\\
     0, & \text{otherwise}
     \end{cases}
\end{equation}

\subsubsection*{Simultaneous truth and performance level estimation (STAPLE) with segmentation quality estimation}
Instead of weighting each segmentation, low quality segmentations with a DSC under a predefined DSC threshold $\epsilon_t \in [0,1]$ are discarded based on their estimated quality and the remaining segmentations with a high estimated DSC are fused with the native STAPLE algorithm implementation \cite{commowick2010incorporating}.

\subsection{Validation} \label{sec:validation}
In our experiments we investigated the following aspects: (1) What is the quality of the proposed segmentation quality estimation (Sec. \ref{sec:qualityPrediction})? (2) What is the quality of the proposed confidence-based approach to segmentation fusion compared to state-of-the-art methods for annotation merging (Sec. \ref{sec:confidenceAnnotationMerging})? (3) How well does the segmentation quality estimation generalize to new object types (Sec. \ref{sec:generalizationCapabilities})? 
and (4) what are the costs of the proposed method compared to state-of-the-art methods (Sec. \ref{sec:annotation_costs})?

\subsubsection{Segmentation quality estimation}
\label{sec:qualityPrediction}
We validated our proposed segmentation concept on a subset of the publicly available PASCAL Visual Object Classes (VOC) \cite{Everingham10}. The regressor was validated on two object classes within the VOC challenge data: $\lbrace{cat, car\rbrace}$. For each class, we used 100 out of the first 150 images, making sure that a broad range of degree of difficulty (e.g. fully visible objects and partially occluded objects) was covered. Using the prototype implementation of our concept (Sec. \ref{sec:concept}) we acquired 10,000 segmentations per class (100 segmentations per image) in a gaming crowd provided by the Pallas Ludens GmbH with their proprietary user interface. Example segmentations with the segmentation outline were provided to the workers. No filters for quality management, e.g. captchas, tutorials or blocking of known spammers, were applied so that clickstreams from a variety of worker types with a high fluctuation in the segmentation quality would be collected. Furthermore we assured that each worker segmented each image at most once per class. This resulted in a total 20,000 segmented images with their corresponding clickstreams for validation.

For each object class we performed a leave-one-out cross validation, thereby ensuring that only annotations of workers that were not involved in the annotations of the training images were considered. To quantify estimation quality, we determined the absolute difference between the true DSC and the estimated DSC of all test data. 

To investigate the number of annotations required for regressor training, we further determined the performance in terms of the $R^2$ score (Eq. \ref{eq:r2score}) as a function of the number of training annotations. Again, we ensured that no worker was included in the test and training data at the same time. For training, we used 100 random permutations of $n$ annotations (for $n$ = 10,000 only one permutation was possible).

Due to the success of wrapper and filter methods for optimal feature selection, we applied these methods for quantifying the relevance of our various features used by the segmentation quality estimation. To avoid bias towards a specific feature selection method, we performed the analysis by applying the most commonly used methods. In particular we employed the wrapper methods \emph{Sequential Forward Selection (SFS)} \cite{whitney1971direct} and \emph{Best First Search (BFS)} \cite{kohavi1997wrappers} with a mean squared error criterion and a feature set size penalty to sequentially build an optimal feature set. Furthermore a wide array of filter methods for feature selection was applied to derive feature sets using the same criterion employed for the wrapper methods. The filter methods include \emph{Conditional Mutual Information Maximization (CMIM)} \cite{fleuret2004fast}, \emph{Interaction Capping (ICAP)} \cite{Jakulin2005}, \emph{Joint Mutual Information (JMI)} \cite{Yang1999}, \emph{Conditional Infomax Feature Extraction (CIFE)} \cite{Lin2006}, and \emph{Mutual Information for Feature Selection (MIFS)} with a nearest neighbor mutual information estimator \cite{battiti1994using,kraskov2004estimating}. These filter methods construct feature sets sequentially, but do not return which number of features must be selected for optimal results. We addressed this shortcoming by running a cross-validation on the training set using the selected feature sets of increasing size and applying the same selection criterion as in the wrapper method for determining the ideal number of features. Feature selection was solely carried out on the training set.

\subsubsection{Confidence-weighted annotation merging} 
\label{sec:confidenceAnnotationMerging}

We compared our method to create confidence-weighted crowd-sourced segmentations (as presented in Sec. \ref{sec:confidence_voting}) with the widely used majority voting method \cite{maierh14Masses} (Sec.  \ref{sec:majorityVoting_SOA}). Furthermore we compared the STAPLE algorithm using raw crowd generated segmentations with the quality estimation enhanced STAPLE algorithm based approach (Sec. \ref{sec:confidence_voting}), where low quality segmentations are filtered out by applying the segmentation quality estimation presented in this paper. To ensure that we only used high quality annotations we executed our method for the DSC at a threshold of $\epsilon_t = 0.9$. According to experiments on a separate data set, this threshold provides a good trade off between quality and excluding high quality annotations.
To investigate the performance of our approach as a function of the number of images, we used $\lambda \in B = \lbrace{ 1,\cdots,10 \rbrace}$ annotations per image that had an equal or higher estimated DSC than $\epsilon_t$. For $\lambda = 1$ the approach basically reduces to the segmentation quality estimation without any further merging of annotations. Considering the fact that poor annotations may be rejected on common crowdsourcing platforms, we compared our method with majority voting using $\lambda$ annotations. In addition, we computed the average number of annotations $\varphi$ required to obtain $\lambda$ annotations with a estimated DSC above $\epsilon_t$ by computing $\varphi = \lambda + r$, where $r$ is the mean number of rejected annotations. For each $\lambda \in B$ our method was compared to majority voting with $\lambda$ and $\varphi$ annotations. Analogously, our STAPLE approach with quality estimation was compared to the native STAPLE algorithm using $\lambda$ and $\varphi$ annotations respectively. The experiments for annotation merging were conducted separately for each class, i.e. training and testing on only cars or cats. 

\subsubsection{Generalization capabilities}
\label{sec:generalizationCapabilities}

To investigate the generalization capabilities of our approach, we applied it to a different crowd and additional validation data using an open re-implementation of the annotation software. Specifically, we re-implemented the annotation concept with  a user interface based on the Openlayers\footnote{https://openlayers.org} \cite{hazzard2011openlayers} library for the widely used crowdsourcing platform Amazon Mechanical Turk (MTurk)\footnote{https://www.mturk.com} \cite{kittur2008crowdsourcing}. The set of object classes obtained from the VOC validation set, namely $\lbrace{car, cat\rbrace}$, was extended by seven further object classes from the COCO data set \cite{lin2014microsoft} yielding the following three object categories: Vehicles $\lbrace{car, airplane, motorcycle\rbrace}$, animals  $\lbrace{cat, bird, dog\rbrace}$, rectangular-shapted objects $\lbrace{laptop, refrigerator, tv\rbrace}$.
For each class, we acquired 1,000 annotations (10 per image; 100 images per class). To investigate how our method degrades for target classes that are further away from the training classes, we determined the performance of segmentation quality estimation when training on the animal and vehicle classes respectively and testing the segmentation quality estimation on (1) the same class, (2) the same category (here: different animals/vehicles), (3) a similar category (here: vehicles for animal classes; animals for vehicle classes) and (4) a different category (here: rectangular-shaped object)).

\subsubsection{Comparison of annotation costs} \label{sec:annotation_costs}
Depending on the number of requested segmentations $a$, the total cost for  $c(a)$ image segmentations can roughly be approximated for the proposed method, a baseline method and the manual grading method (Sec. \ref{sec:manualgraqding_SOA}) applied in \cite{lin2014microsoft}.

\subsubsection*{Proposed method}
The total costs to acquire $a$ segmentations with our proposed method can be approximated as follows, taking into account the average number of annotations used to perform the confidence-weighted majority voting step $a_{mv}$, the number of annotations used to train the regressor $a_t$ and the percentage of spam $s$ to be expected:
\begin{equation}
c(a) = a_t+\left(a+\frac{s}{1-s}a\right) \cdot a_{mv}
\end{equation}

\subsubsection*{Baseline method}
A widely used method for simple quality control in crowd based annotation merging using majority voting is to let each crowd worker annotate an image with known reference annotation every $a_w$ annotations. For $a_{mv}$ annotations needed for majority voting typically 10-30\% of quality control tasks are mixed in between the tasks \cite{bragg2016optimal}. The total cost of annotating $a$ images can then be calculated as follows, also taking into account the number of reference annotations $a_r$:
\begin{equation}
c(a) = a_{mv}a + \frac{a_{mv}a}{a_w-1} + a_r
\end{equation}


\subsubsection*{Manual grading of annotation quality}
A general estimation of the annotation costs entailed by the method applied in \cite{lin2014microsoft} can be determined by considering the average number of annotations performed by one worker $A_{aw}$, the number of categories $n_c$ annotations are requested for, the total number of recruited workers $n_w$, the total number of approved workers $n_{aw}$ to create $c(a)$ annotations, the number of banned approved workers $r$ and the costs associated with the verification stage $v$ for a percentage of spam $s$:
 
\begin{equation}
c(a) = \frac{a}{1-s} + n_c n_w + A_{aw} n_{aw} \cdot r + v 
\end{equation}



\section{Results} \label{sec:results} 

\subsection{Segmentation quality estimation} \label{sec:resQualityPrediction}
\begin{figure}[!!!htb]
  \centering
  \subfloat[]{\label{fig:prediction-error}\includegraphics[width=0.54\linewidth]{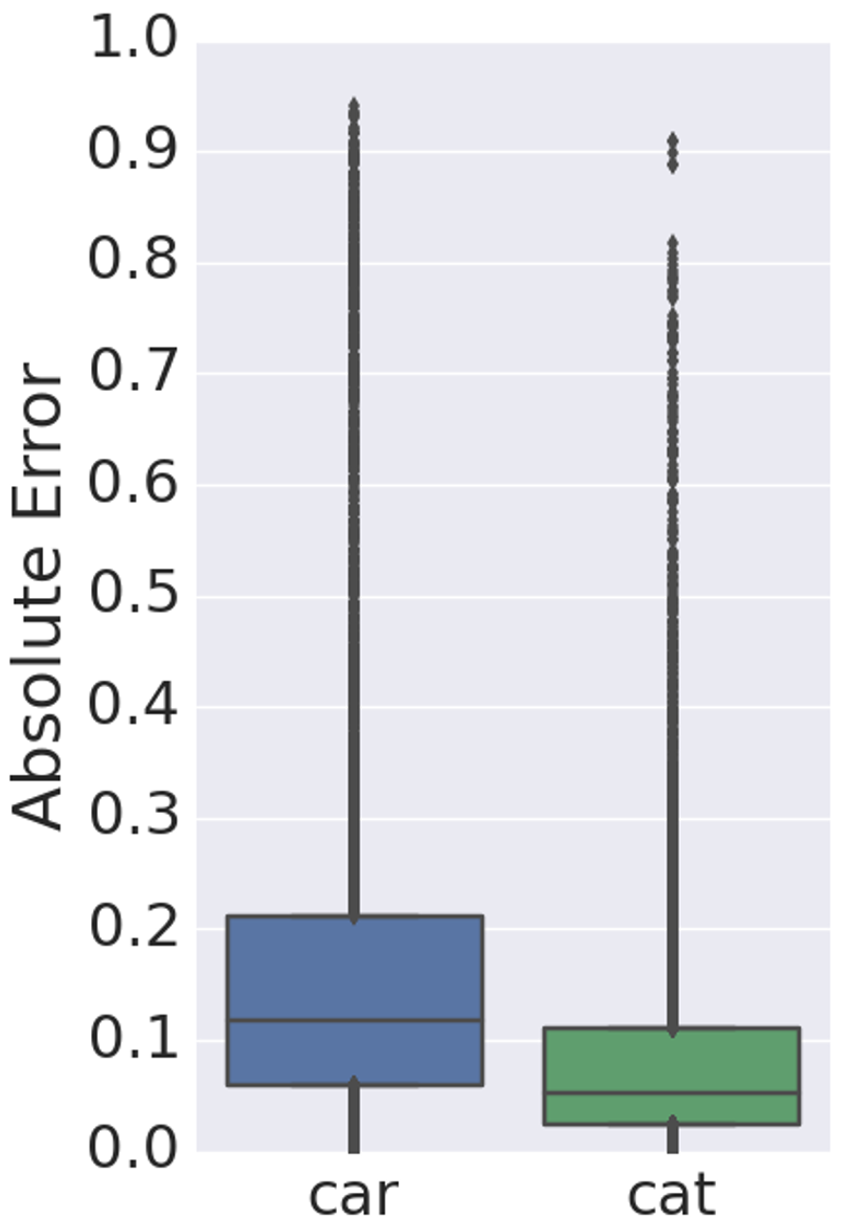}}
  \subfloat[]{\label{fig:bad-predictions-piecharts}\includegraphics[width=0.46\linewidth]{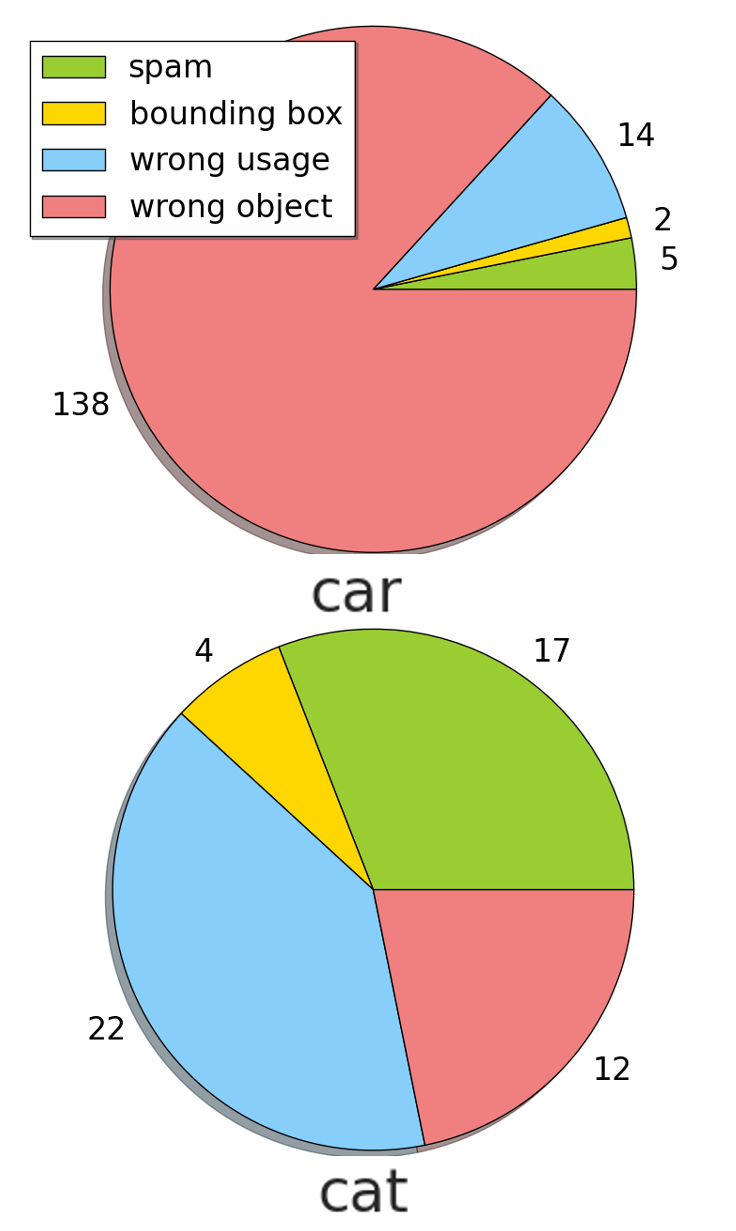}}
  \caption{(a) Absolute error of the estimated segmentation quality for training and testing on the same class. (b) Distribution of crowd segmentations that were estimated to have a high DSC but had a low true DSC (false positives) divided into the error classes introduced in Sec. \ref{sec:resQualityPrediction} with the absolute amount for each error class. The total amounts of false positives relative to all estimations for each class were: $3\%$ (car) and $1\%$ (cat).}
\end{figure}

By filtering the segmentations with our segmentation quality estimation method (using $\epsilon_t = 0.9$, $\lambda=3$) we improve the mean, median (inter quartile range) (IQR) quality of the pool of segmentations by 16\%, 4\% (IQR:18\%,3\%) from 0.80, 0.91 (IQR: 0.79,0.94) to 0.93, 0.95 (IQR: 0.93,0.97).
The mean and median (IQR) absolute difference between the true DSC and the estimated DSC were $0.18$, $0.12$ (IQR: $0.06$, $0.21$) for training and testing on cars and $0.09$, $0.05$ (IQR: $0.02$, $0.11$) for training and testing on cats (Fig. \ref{fig:prediction-error}).

\begin{figure}[!!!htb]
  \centering
  \includegraphics[width=\columnwidth]{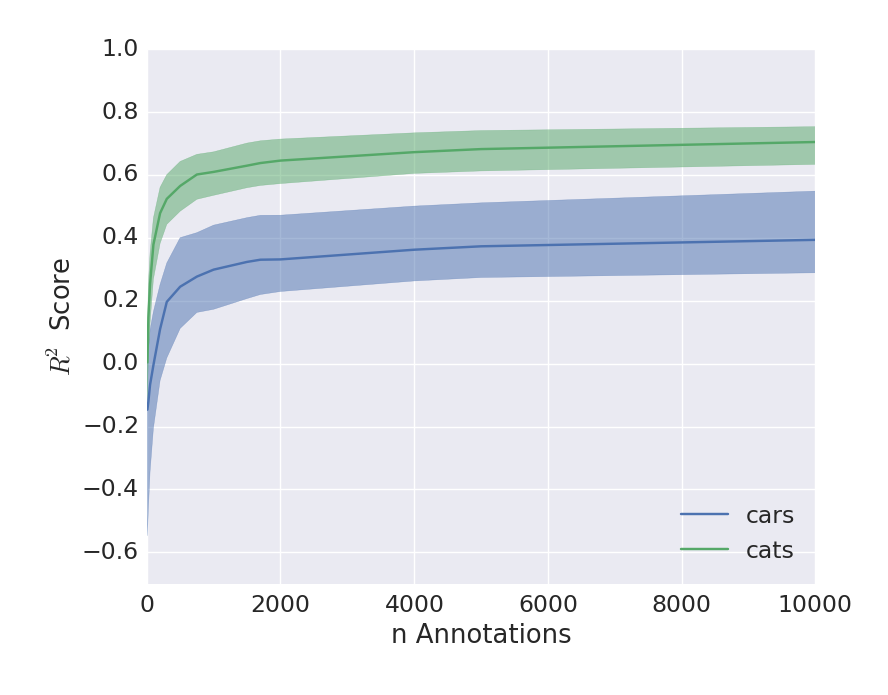}
    \caption{Median $R^2$ score and IQR as a function of the number of images used to train the segmentation quality estimation.}
    \label{fig:TrainingSetSize}
\end{figure}

In order to create high quality annotations, it is particularly crucial to omit low quality segmentations. To assess the performance of our method in view of this crucial aspect, we identified all annotations with a poor true DSC below 0.5 and a good estimated DSC over 0.8. These false positive (FP) estimations made up $2\%$ of all annotations used for validation. They are subdivided into the following four categories, as illustrated in Fig. \ref{fig:bad-predictions-examples}:
\begin{figure}[!!!htb]
  \captionsetup[subfigure]{labelformat=empty}
  \centering
  \subfloat[]{\label{fig:bad-predictions-example-wrong-object-baby}\includegraphics[width=0.238\linewidth]{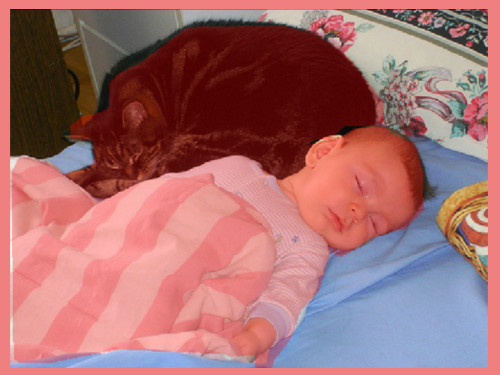}}
  \hspace{0.05cm}
  \subfloat[]{\label{fig:bad-predictions-example-wrong-usage-outline}\includegraphics[width=0.235\linewidth]{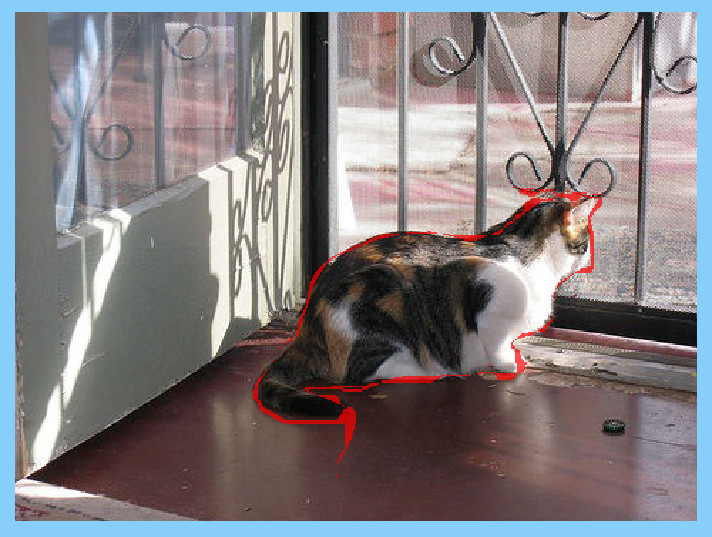}}
  \hspace{0.05cm}
  \subfloat[]{\label{fig:bad-predictions-example-spam-cat}\includegraphics[width=0.1752\linewidth]{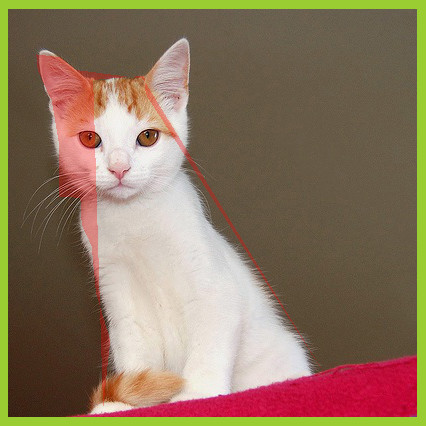}}
  \hspace{0.05cm}
  \subfloat[]{\label{fig:bad-predictions-example-bounding-box-cat}\includegraphics[width=0.266\linewidth]{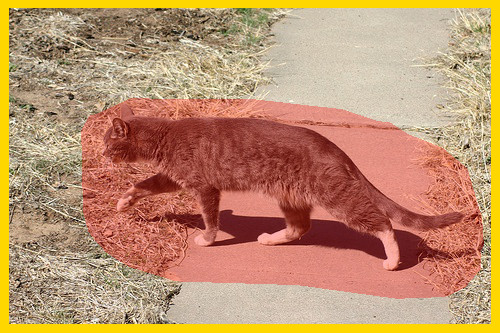}}
  \vspace{-1cm}
  
  \subfloat[(a)]{\label{fig:bad-predictions-example-wrong-object-bike}\includegraphics[width=0.238\linewidth]{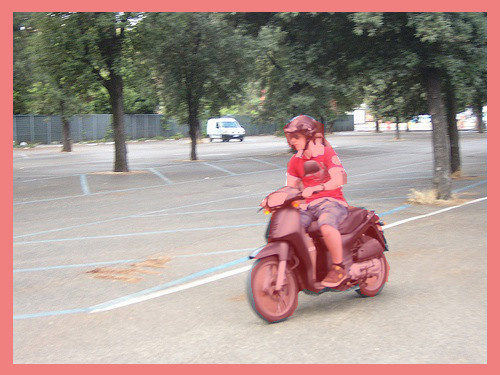}}
  \hspace{0.05cm}
  \subfloat[(b)]{\label{fig:bad-predictions-example-wrong-usage-inverted}\includegraphics[width=0.235\linewidth]{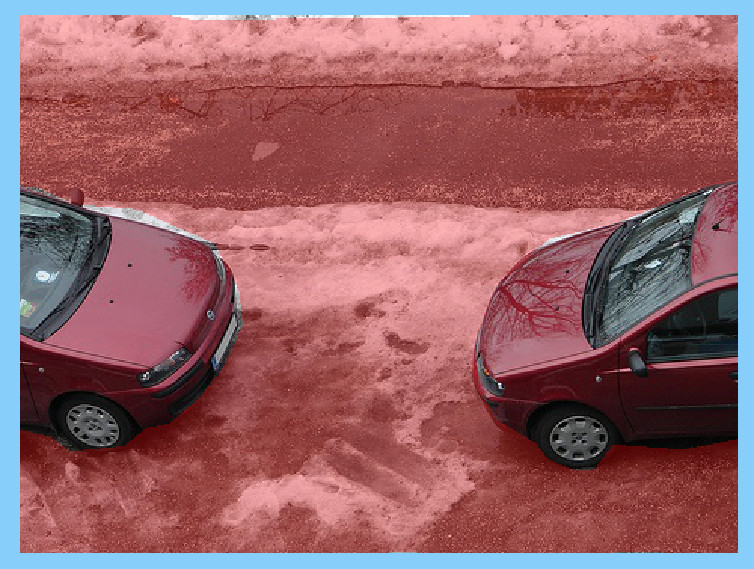}}
  \hspace{0.05cm}
  \subfloat[(c)]{\label{fig:bad-predictions-example-spam-car}\includegraphics[width=0.1752\linewidth]{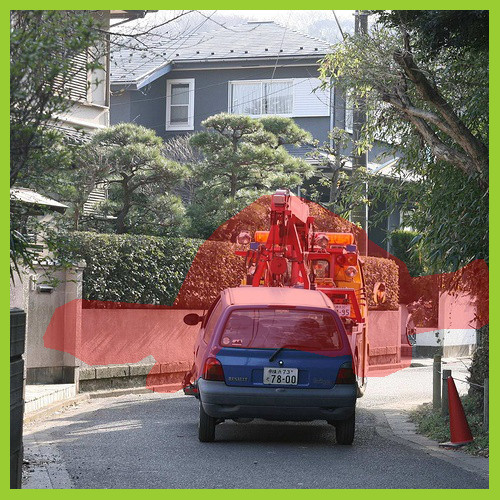}}
  \hspace{0.05cm}
  \subfloat[(d)]{\label{fig:bad-predictions-example-bounding-box}\includegraphics[width=0.266\linewidth]{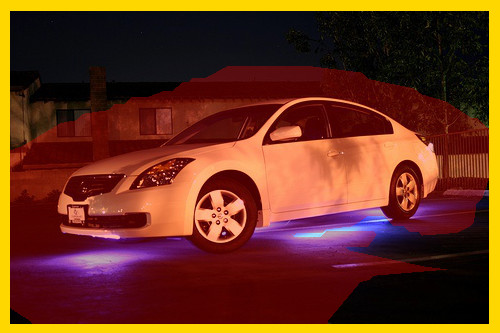}}
  
  \caption{Examples of poor estimations: (a) Accurate segmentation of wrong objects. (b) Wrong usage of the segmentation tool: Crowd users draw outlines with polygons (top). Inverted segmentation of the object (bottom). (c) Spam. (d) Bounding box.}
    \label{fig:bad-predictions-examples}
\end{figure}
(a) Wrong object: The worker creates an accurate segmentation of a wrong object (possibly on purpose). (b) Wrong tool usage: Wrong usage of the annotation tool, e.g. inverted segmentation or workers try to draw outlines with polygons rather than covering the objects with a polygon. (c) Spam: Obvious spam. (d) Bounding box: Workers draw a bounding box around the object of interest rather than an accurate contour. The distribution of false positive quality estimations according to these error classes is shown in Fig. \ref{fig:bad-predictions-piecharts}.
The segmentation quality as a function of the number of images used to train the regressor is displayed in Fig. \ref{fig:TrainingSetSize}.
For both data sets (cat and car), \emph{the mean angle between image gradient direction and interpolated vertex normal} and the \emph{ratio of traveled mouse distance to length of the segmented contour} were found to be important features by all feature selection methods. A potentially high impact on the estimation accuracy could also be found for the \emph{median mouse velocity}, the \emph{median mouse velocity for draw events}, \emph{the number of events in the clickstream}, and the \emph{median angle between image gradient direction and mouse move direction for draw events} features. The best feature selection methods found feature sets as small as  6 features that achieved the same performance in the test set as we obtained when using the full feature set (TABLE \ref{tab:feature_selection}). Of note, when using only image-based features that are calculated on the result and do not rely on any additional clickstream information, the mean error is approximately twice as high for both data sets compared to the errors reported in TABLE \ref{tab:feature_selection}.
\begin{table}[t]
    \centering
\begin{tabular}{l|c|r|r}
\hline
data set & method & no. features & mean error\\\hline\hline
cat & CMIM & 7 & 0.07\\\hline
cat & ICAP & 33 & 0.06\\\hline
cat & JMI & 11 & 0.06\\\hline
cat & CIFE & 6 & 0.06\\\hline
cat & MIFS & 7 & 0.06\\\hline
cat & SFS & 8 & 0.07\\\hline
cat & BFS & 7 & 0.07\\\hline
cat & BASE & 53 & 0.06\\\hline\hline
car & CMIM & 11 & 0.10\\\hline
car & ICAP & 6 & 0.10\\\hline
car & JMI & 7 & 0.11\\\hline
car & CIFE & 6 & 0.10\\\hline
car & MIFS & 21 & 0.11\\\hline
car & SFS & 7 & 0.11\\\hline
car & BFS & 8 & 0.11\\\hline
car & BASE & 53 & 0.10\\\hline
\hline
\end{tabular}
\caption{Mean estimation error for each feature selection method. The minimal chosen feature set achieved a similar classification performance compared to all features (BASE).}
\label{tab:feature_selection}
\end{table}

\begin{figure*}[!!!htb]
  \centering
  \subfloat[car]{\label{fig:confidence-voting-cars-cars}\includegraphics[width=0.48\linewidth]{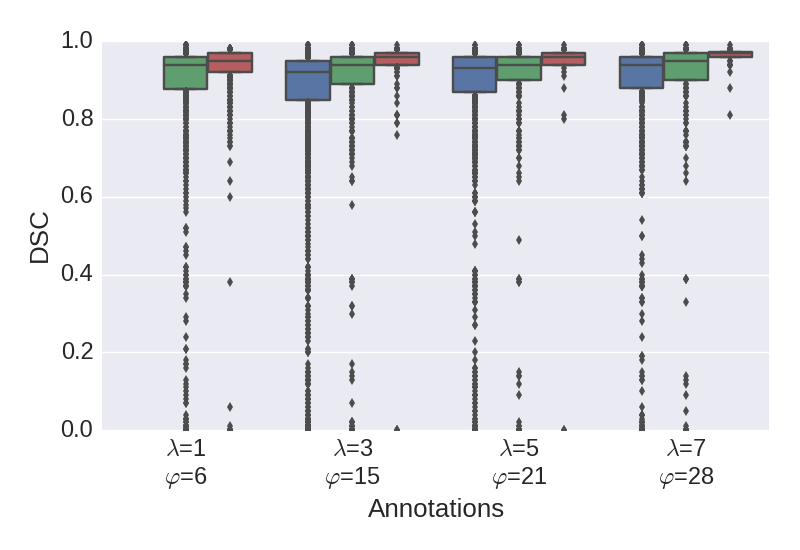}} 
  \subfloat[cat]{\label{fig:confidence-voting-cats-cats}\includegraphics[width=0.48\linewidth]{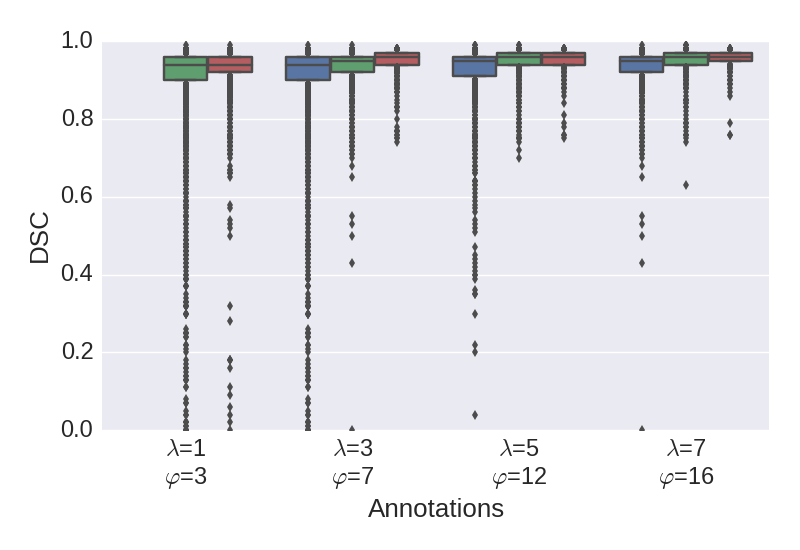}}
  \caption{Confidence-weighted majority voting (red) compared to conventional majority voting with $\lambda$ annotations (blue) and $\varphi$ annotations (green) for different training and testing classes (car and/or cat), where $\varphi$ represents the average number of annotations to obtain $\lambda$ annotations with a estimated DSC above $\epsilon_t$. Performance is assessed for a estimated DSC threshold of $\epsilon_t = 0.9$ and a varying number of annotations $\lambda$. For clarity only subsets of the experiments ($\lambda \in \lbrace{ 1,3,5,7 \rbrace}$) are visualized.}
    \label{fig:confidence-voting-results}
\end{figure*}

\begin{figure*}[!!!htb]
  \centering
  \subfloat[car]{\label{fig:staple-voting-cars-cars}\includegraphics[width=0.48\linewidth]{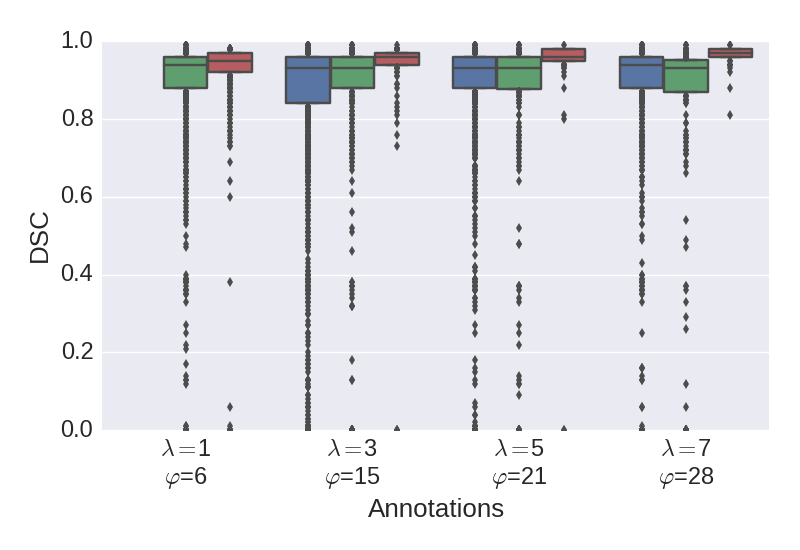}} 
  \subfloat[cat]{\label{fig:staple-voting-cats-cats}\includegraphics[width=0.48\linewidth]{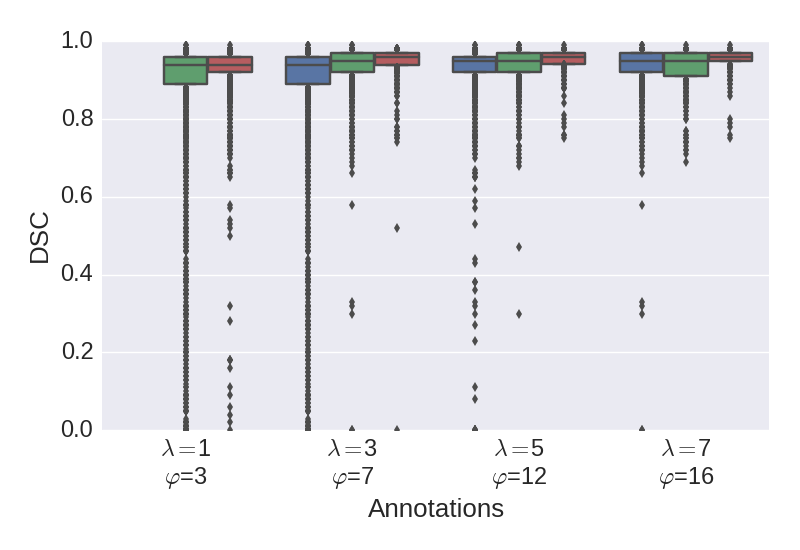}}
  \caption{STAPLE with segmentation quality estimation (red) compared to conventional STAPLE with $\lambda$ annotations (blue) and $\varphi$ annotations (green) for different training and testing classes (car and/or cat), where $\varphi$ represents the average number of annotations to obtain $\lambda$ annotations with a estimated DSC above $\epsilon_t$. Performance is assessed for a estimated DSC threshold of $\epsilon_t = 0.9$ and a varying number of annotations $\lambda$. For clarity only subsets of the experiments ($\lambda \in \lbrace{ 1,3,5,7 \rbrace}$) are visualized.}
    \label{fig:staple-voting-results}
\end{figure*}.

\subsection{Confidence-weighted annotation merging} \label{sec:resConfidenceAnnotationMerging}
The resulting DSC values for different numbers of annotations $\lambda$ for the two confidence-based segmentation merging approaches and the number of rejected crowd segmentations are shown in Fig. \ref{fig:confidence-voting-results} and Fig. \ref{fig:staple-voting-results}. Both approaches using the segmentation quality estimation outperformed their baseline methods in terms of the segmentation quality and were robust to outliers.

Our presented approach for confidence-weighted majority voting (Sec. \ref{sec:confidence_voting}) produced statistically significant better results compared to conventional majority voting for the same amount of annotations (Fig. \ref{fig:confidence-voting-results}). To obtain a median DSC of $0.95$ with our method, the number of required annotations ranged from $1$ to $3$ for the two experiments depicted in Fig. \ref{fig:prediction-error}. Compared to conventional  majority  voting (Majority Voting $\lambda$ in Fig. \ref{fig:confidence-voting-results}), the number of annotations was reduced by $75\%$ on average.

When we used the STAPLE algorithm in conjunction with our segmentation quality estimation (Sec. \ref{sec:confidence_voting}) statistically significant better results were produced compared to the native STAPLE algorithm for the same amount of annotations (Fig. \ref{fig:staple-voting-results}). To obtain a median DSC of $0.95$ with our method, the number of required annotations ranged from $1$ to $2$ for the two experiments depicted in Fig. \ref{fig:prediction-error}. 
Compared to the native STAPLE algorithm (STAPLE $\lambda$ in Fig. \ref{fig:staple-voting-results}), the number of annotations was reduced by $73\%$ on average.

Mean differences between conventional majority voting $\lambda$ and confidence-weighted majority voting and mean differences between the STAPLE algorithm $\lambda$ and our STAPLE approach with segmentation quality estimation both ranged from 0.02 (bootstrapped $95\%$-confidence interval: 0.01, 0.02; 10 annotations, cat) to 0.13 (0.1, 0.16; 4 annotations, car). Non parametric Mann-Whitney U tests for all comparisons with $\lambda \in \lbrace 1,\cdots, 10 \rbrace$ yielded $p$ values that were statistically significant at the significance level of 0.0001, even after a conservative adjustment for multiple testing by the Bonferroni method.
\subsection{Generalization capabilities}
\label{sec:resGeneralizationCapabilities}
\begin{figure}[!!!htb]
  \centering
  \includegraphics[width=\columnwidth]{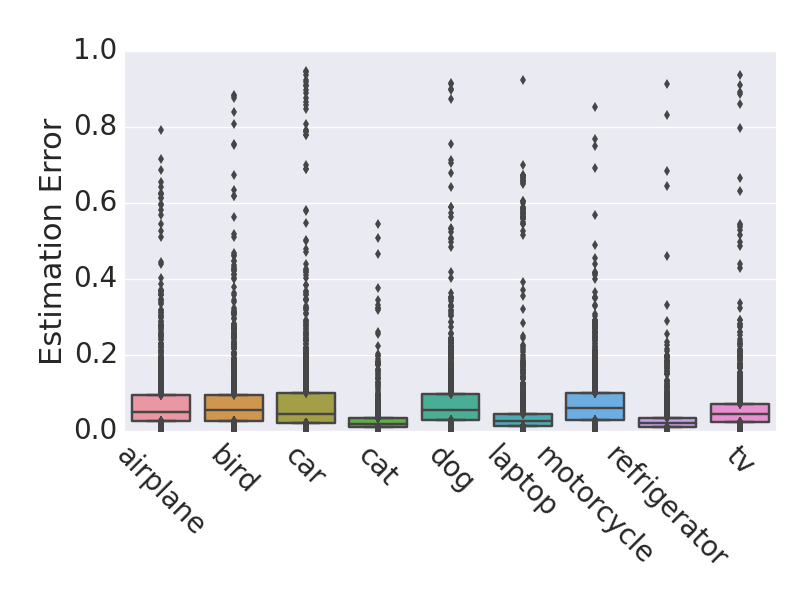}
    \caption{Intra-class segmentation quality estimation performance for all classes (see Sec. \ref{sec:resGeneralizationCapabilities}).}
    \label{fig:EstimationOverview}
\end{figure}

\begin{figure}[!!!htb]
  \centering
  \includegraphics[width=\columnwidth]{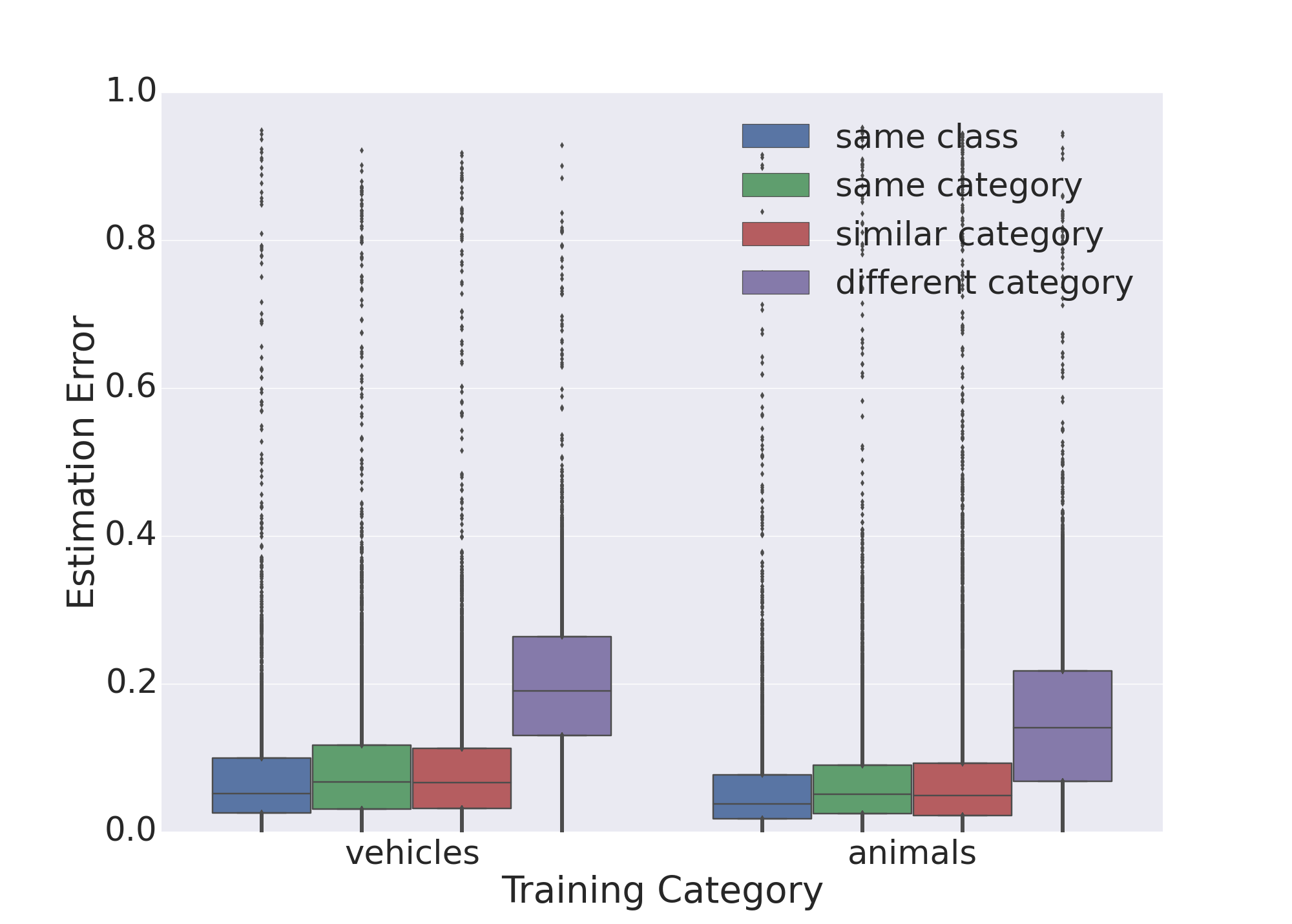}
    \caption{Descriptive statistics of the quality estimation error when training on animals or vehicles and testing on (1) the same class, (2) the same category (here: different animals or vehicles), (3) a similar category (here: vehicles or animals) and (4) a different category (here: rectangular-shaped objects).}
    \label{fig:PerformanceDegradation}
\end{figure}

When applied to another crowd with additional validation data, intra-class performance of our segmentation quality estimation remains high, even when less than 1,000 annotations are used for training (Fig. \ref{fig:EstimationOverview}). As shown in Fig. \ref{fig:PerformanceDegradation}, the estimation performance degrades the further the target class is away from the training class.  Yet, even when training on an animal class and testing on a vehicle (or vice versa), the mean/median estimation error was still below 0.1 and the number of annotations compared to conventional majority voting can be reduced by 50 \%.

\subsection{Comparison of segmentation costs} \label{sec:res_annotation_costs}
\begin{figure}[!!!htb]
  \centering
  \includegraphics[width=\columnwidth]{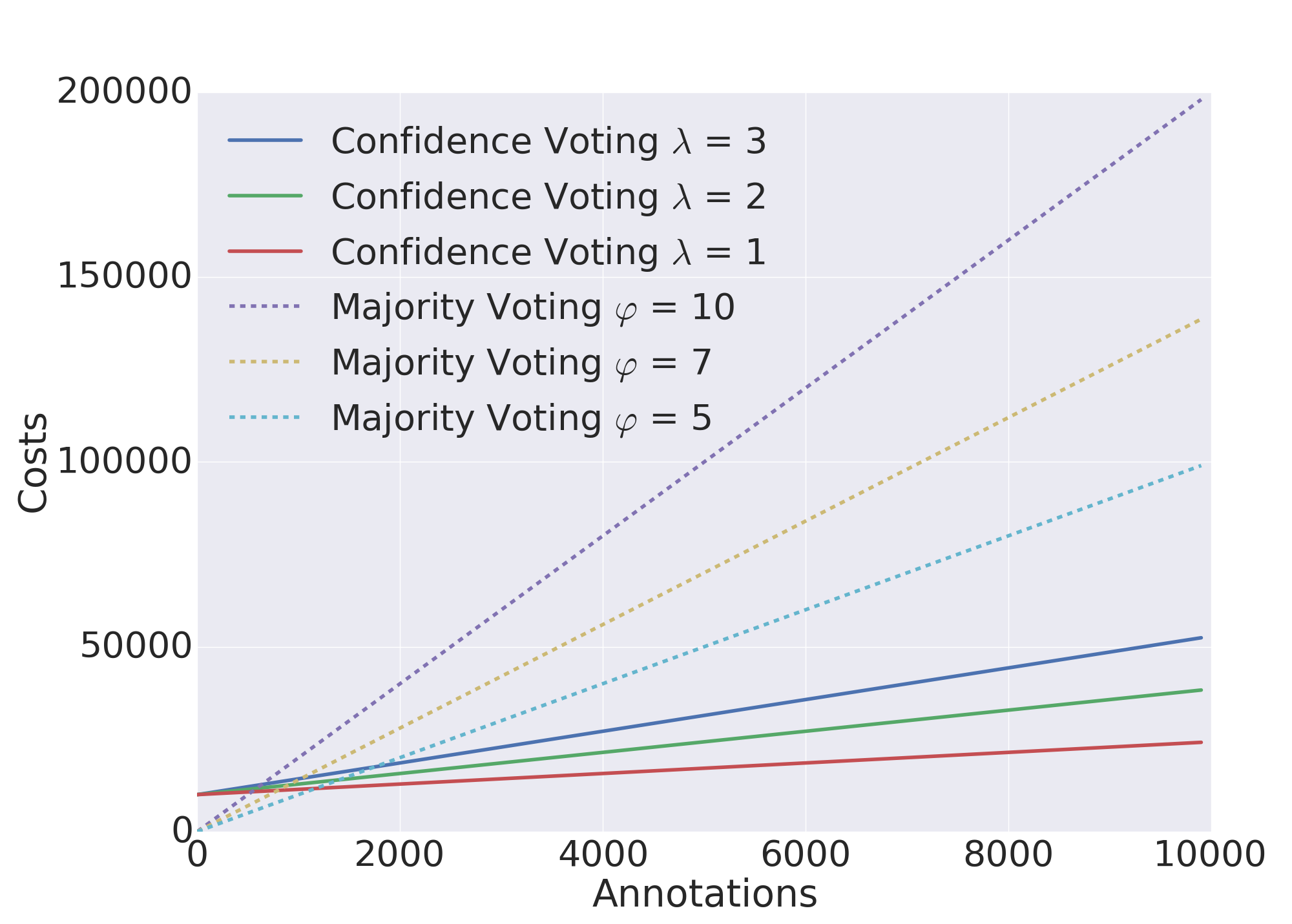}
    \caption{Comparison of the annotation costs of the proposed method for different $\lambda$ with a baseline method based on majority voting for different $\varphi$ with $s=30\%$ of spam, $a_t=10,000$ training annotations for the proposed method, and $a_w=10$ quality control tasks for the baseline method. The annotation costs are plotted for the number of requested annotations.}
    \label{fig:ConfidenceToBaseline}
\end{figure}

\begin{figure}[!!!htb]
  \centering
  \includegraphics[width=\columnwidth]{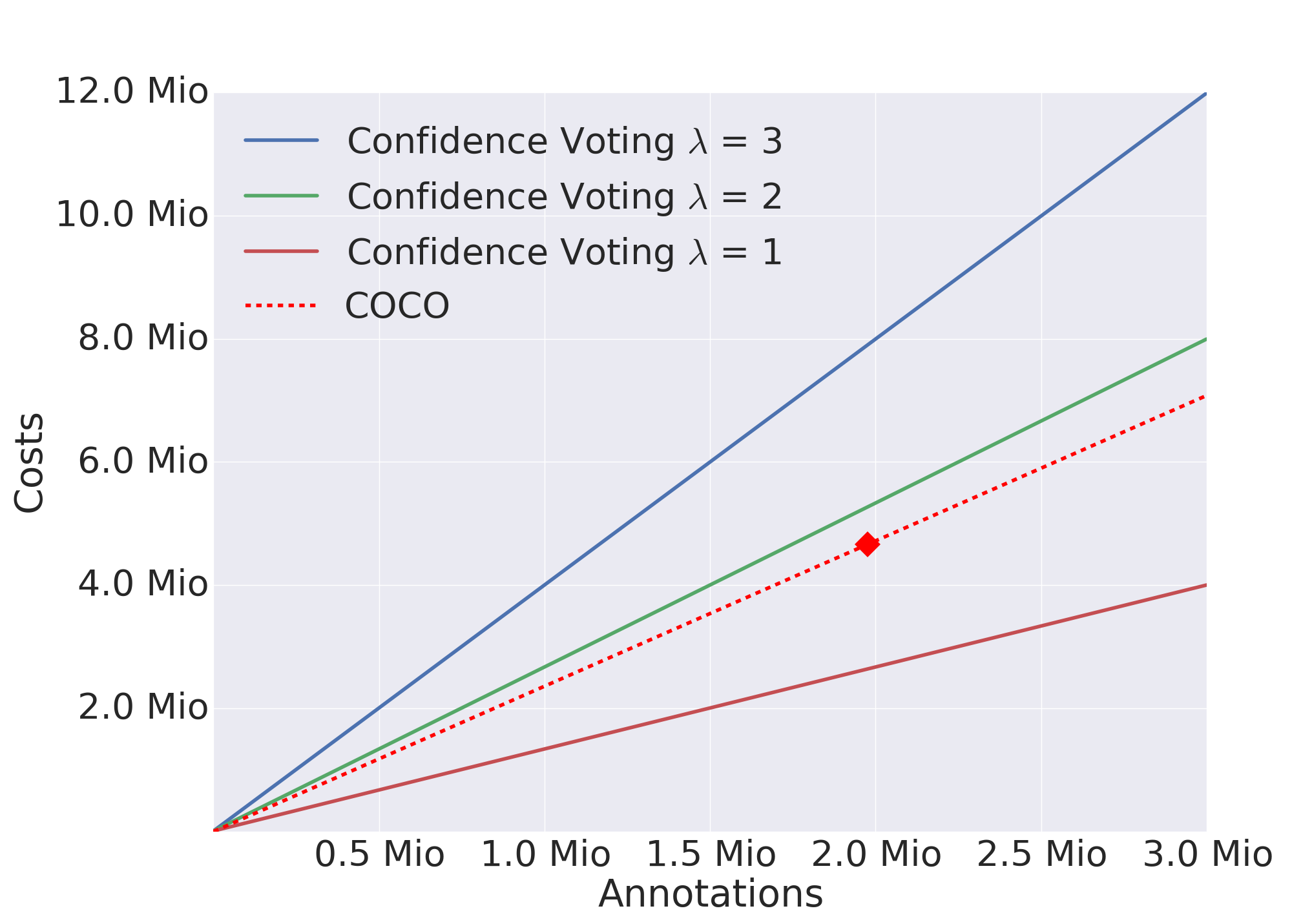}
    \caption{Comparison of the annotation costs of the proposed method for different $\lambda$ with the estimated costs of the approach applied by the method of Lin et al. \cite{lin2014microsoft} (COCO) with $s=24.9\%$ of spam. The annotation costs are plotted for the number of segmentations requested. The diamond represents the estimated total cost for annotating the full COCO data set.}
    \label{fig:ConfidenceToCoco}
\end{figure}

Annotation costs of our method compared to majority voting and the manual grading method are shown in Fig.  \ref{fig:ConfidenceToBaseline} and Fig. \ref{fig:ConfidenceToCoco}, respectively. 

It can be seen that our method outperforms the baseline method (majority voting) in terms of costs when a typical number of $>1,000$ annotations is acquired. Our approximations further indicate that the manual grading method for contour drawing is more expensive than our method instantiated with $\lambda = 1$ and less expensive for $\lambda \geq 2$ when assuming the percentage of spam to be $\sim$25\%. Note that the costs $v$ for verification required by manual grading were not considered in our analysis (as the numbers were not available). As our method may potentially be combined with a verification step as well, instantiation with $\lambda = 1$ would be feasible. 

\section{Discussion} 
\label{sec:discussion}
To our knowledge, we have presented the first approach to crowd-based object segmentation that uses annotation process-based features derived from clickstreams for quality control. In contrast to previous approaches, this enables us to estimate segmentation quality without using any prior knowledge of specific workers and without having to perform any additional tasks depending on known reference data once the segmentation quality estimation is trained. 
Our work was inspired by user behavior analysis with clickstreams  \cite{wang2016unsupervised,ting2005ubb,srivastava2000web,gunduz2003web} which has already successfully been explored outside the field of crowdsouring for e-commerce applications \cite{su2015method}, social networks \cite{sybils} and web browsing behavior analysis \cite{heer2002separating,obendorf2007web}. 
 
Our experiments support the following three hypotheses:

\begin{enumerate}
\item Clickstream features are very good predictors for segmentation quality
\item These features generalize over (similar) object classes
\item Clickstream-based annotation quality estimation can be applied for confidence-based annotation merging 
\end{enumerate}
 
In a scenario where bad quality segmentations can be directly rejected without rewarding the workers, we were able to achieve the same segmentation accuracies while reducing the crowd annotation costs by up to $75\%$ compared to conventional methods. Even when using only a single annotation through our segmentation quality estimation we were already able to outperform the baseline methods in terms of segmentation quality. Importantly, our experiments related to feature selection suggest that the annotation process-based features are crucial for the success of our method. In addition, our presented method is resilient to outliers by rejecting bad segmentations through the quality estimation step and weighting them based on their estimated quality, unlike to conventional majority voting and the STAPLE algorithm. We can also show that our regressor does not need to be trained on the object class it is applied to, but generalizes well across classes, rendering the costs required for training the regressor negligible. According to initial experiments presented in Sec. \ref{sec:resGeneralizationCapabilities} the training class should be chosen as closely as possible to the testing class. 

To create highly accurate object segmentations it is crucial that inaccurate annotations are detected with a high reliability. The majority of estimations with incorrect high quality estimations could be traced back to accurate segmentations of the wrong object on the car data set (Fig. \ref{fig:bad-predictions-examples}a). This occurred particularly frequently on rather difficult tasks, e.g. where the car is hidden in the background (Fig \ref{fig:bad-predictions-examples}a) or partially occluded and the workers tend to segment foreground objects like pedestrians, animals or motorcycles in addition to or instead of the object of interest. In contrast to the cat data set, where the object of interest is mostly visible in the foreground, some images of cars were taken in an environment with traffic showing a higher amount of different objects. For example trucks and busses are considered as a different class, but pick-up trucks and mini vans belong to the class car. This can be misleading for some workers and in some images cars and trucks are hard to distinguish from each other. In this case the workers tend to create accurate segmentations of both vehicles, while only the segmentation of the car is included in the reference data. In contrast to the car data set, the cat data set does not suffer from particularly difficult tasks, but has a slightly higher amount of misclassified bounding box segmentations and spam. With the low overall estimation error (Fig. \ref{fig:prediction-error}), the amount of misclassified spam can be considered as negligible. The largest proportion of erroneous classifications on the images of cats was produced by workers that used the segmentation tool in a wrong way, e.g. inverted segmentations (Fig. \ref{fig:bad-predictions-examples}b), drawing outlines with polygons. It should be pointed out that all of these problems are related to issues with the annotator instructions rather than to the quality estimation method itself. Of note, the method was designed for estimating the quality of a single object segmentation. It can easily be extended with instance spotting step to point the worker to the object of interest and assure the segmentation the correct object, as presented in \cite{lin2014microsoft}. Clear task instructions and training workers with tutorials might also help to minimize further problems \cite{gottlieb2012pushing}.  

We are currently also not considering the history of a specific worker. It is likely that a worker who has provided low-quality annotations several times will not be a reliable worker for future tasks. A pre-selection of workers could be performed in this case to achieve the desired segmentation result more quickly. We opted not to perform additional steps for quality assurance to better validate our method. Furthermore, spammers trying to cheat the system can create new accounts as soon they are blocked and will continue spamming the system until they are identified. Using a similar mouse dynamic based features set classification approach as described in Feher et al.\cite{Feher2012} for user identity verification and taking into account the work for clickstream based sybil account detection in online communities \cite{sybils} and user behaviour clustering \cite{wang2016unsupervised} our concept could be extended to detect known malicious workers on different accounts. 

The presented segmentation concept is rather general and open in the way it can be implemented. For this paper we chose standard and widely used methods for the individual components. We believe that more sophisticated approaches for annotation merging, like the maximum aposteriori STAPLE algorithm approach \cite{commowick2012estimating} for merging single object annotations could further improve segmentation quality.

Given that the chosen features only rely on the clickstream data and gradient information extracted from the image and that no absolute pixel values are considered, we believe that our method can be applied to a wider range of different domains such as bio-medical imaging without needing to retrain the classifier. In view of the lack of publicly available reference data in the bio-medical imaging domain \cite{greenspan2016guest}, this could result in a huge benefit for data annotation in this field. Additionally, future work will investigate adapting our method to other classes of annotation, such as localization using bounding boxes. Finally, methods for crowd-algorithm collaboration could be investigated (e.g. \cite{maier2016crowd,carlier2014click}) to further reduce annotation costs. In conclusion, we believe that our method has a great potential for use in large-scale low-cost data annotation.

\section*{Acknowledgements}
The authors would like to thank Caroline Feldmann for her help with the figures and Pallas Ludens GmbH for supporting us with their crowdsourcing platform.
This work has been financed by the Klaus Tschira Foundation (project: "Endoskopie meets Informatik $-$ Pr\"azise Navigation f\"ur die minimalinvasive Chirurgie") and SFB/TRR 125 - Cognition-Guided Surgery.
Finally, the authors would like to thank Esther Stenau, Sebastian Wirkert, Caroline Feldmann, Janek Gr\"ohl, Annika Reinke, Hellena Hempe, Sarina Thomas, Clemens Hentschke and Matthias Baumhauer for helping with the data acquisition as well as Tsung-Yi Lin from Cornell Tech, Cornell University for providing further statistics on the COCO data set.





\ifCLASSOPTIONcaptionsoff
 \newpage
\fi




\bibliographystyle{ieeetr}
\bibliography{egbib}

%

\begin{IEEEbiography}[{\includegraphics[width=1in,height=1.25in,clip,keepaspectratio]{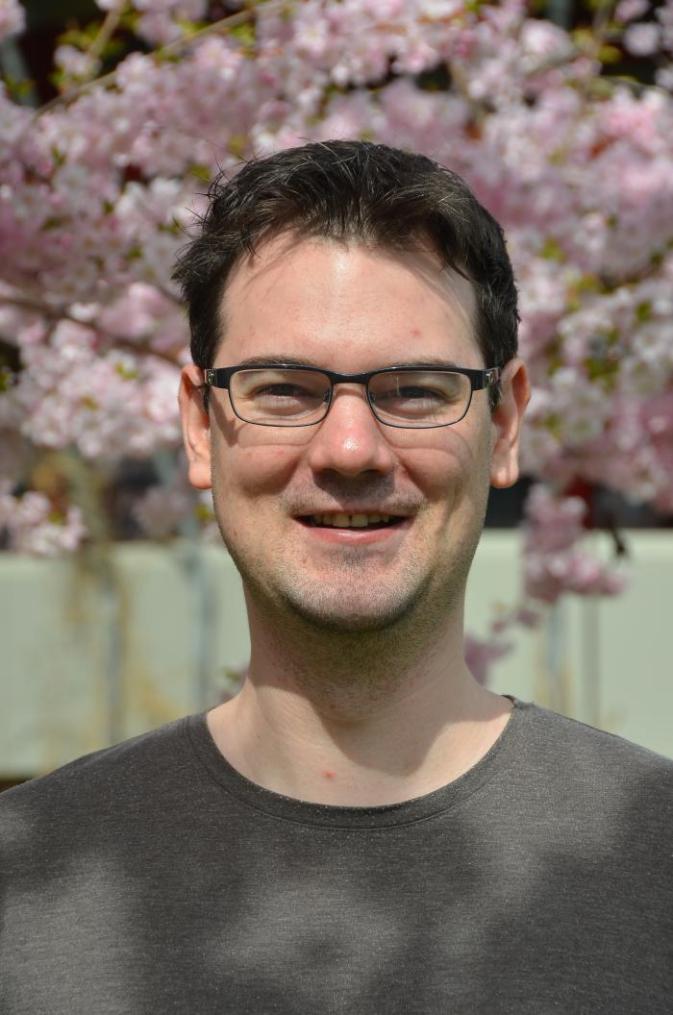}}]{Eric Heim}
received his MSc in Applied Computer Science from the University of Heidelberg in 2013. He is currently working in the Division of Computer Assisted Medical Interventions at the German Cancer Research Center (DKFZ) in Heidelberg towards his PhD degree at the Interdisciplinary Center for Scientific Computing (IWR) University of Heidelberg. His research focuses on crowdsourcing enhanced algorithms in the context of medical procedures.
\end{IEEEbiography}

\begin{IEEEbiography}[{\includegraphics[width=1in,height=1.25in,clip,keepaspectratio]{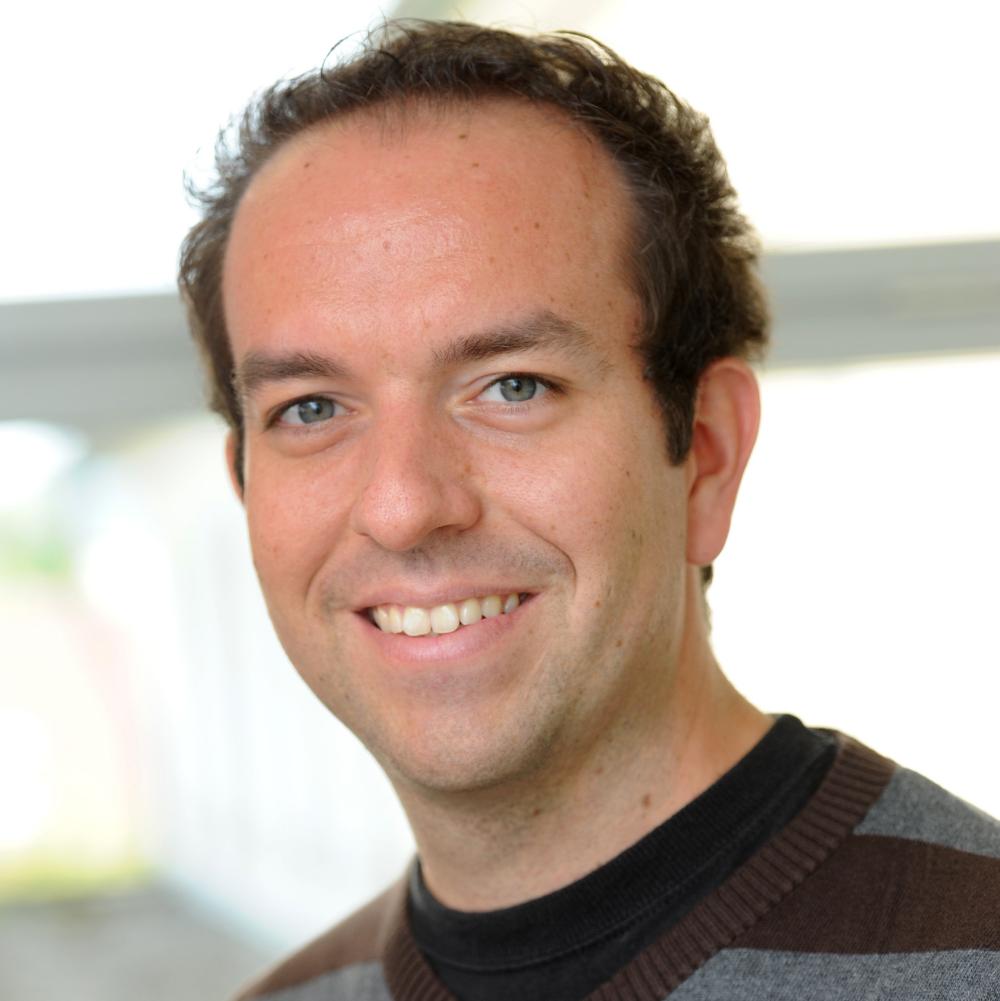}}]{Alexander Seitel} received his Doctorate in Medical Informatics from DKFZ / University of Heidelberg in 2012 and holds a Diploma degree in Computer Science from the Karlsruhe Institute of Technology (2007). He is currently a postdoctoral researcher at the German Cancer Research Center (DKFZ) focussing on computer-assisted interventions in particular navigation solutions for percutaneous needle insertions. 
\end{IEEEbiography}

\begin{IEEEbiography}[{\includegraphics[width=1in,height=1.25in,clip,keepaspectratio]{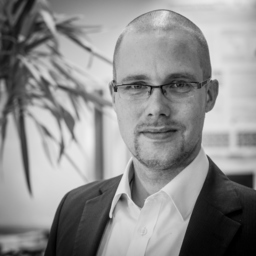}}]{Jonas Andrulis}
received a MSc in Economics Engineering from Karlsruhe Institute of Technology KIT in 2008 specialising in statistical modeling. In his masters thesis he evaluated counterparty reputation risk in a multi-layer Bayesian network. Since then he increasingly investigated user-incorporating models to support ground truth generation with uncertainty.
\end{IEEEbiography}

\begin{IEEEbiography}[{\includegraphics[width=1in,height=1.25in,clip,keepaspectratio]{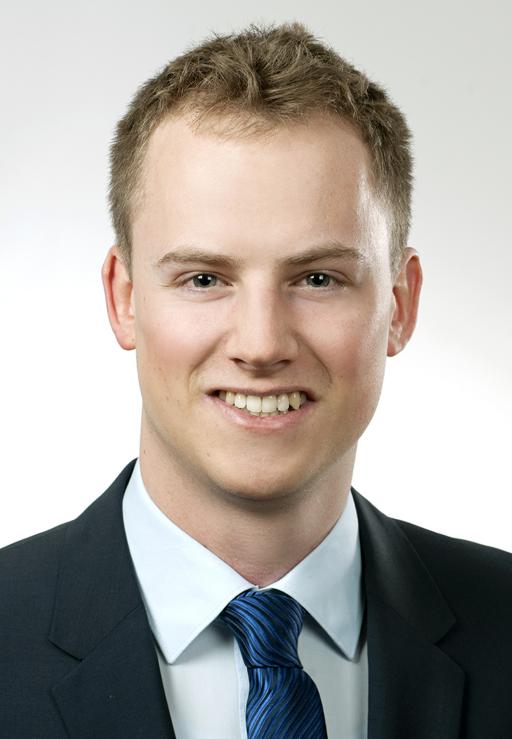}}]{Fabian Isensee}
holds a MSc in Molecular Biotechnology from the University of Heidelberg. In his masters thesis he investigated feature selection and feature importance in the context of image segmentation. He is currently working as a PhD student in the Division of Medical Image Computing at the German Cancer Research Center (DKFZ) where his research focuses on the development of deep learning algorithms for medical image classification and segmentation.
\end{IEEEbiography}

\begin{IEEEbiography}[{\includegraphics[width=1in,height=1.25in,clip,keepaspectratio]{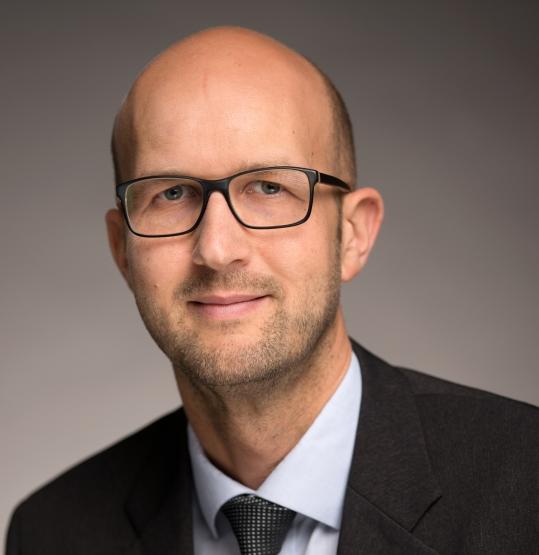}}]{Christian Stock} holds an MSc in Biostatistics from Heidelberg University, a MSc in Health Services Research from York University (UK), and a Doctorate from Heidelberg University. He is a senior statistician and epidemiologist at the Divison of Clinical Epidemiology and Aging Research at the German Cancer Research Center (DKFZ) and at the Institute of Medical Biometry and Informatics at Heidelberg University Hospital. His work focuses on the development of statistical models for complex observational data and on applied statistics for program and method evaluation studies in the bio-medical field.
\end{IEEEbiography}

\begin{IEEEbiography}[{\includegraphics[width=1in,height=1.25in,clip,keepaspectratio]{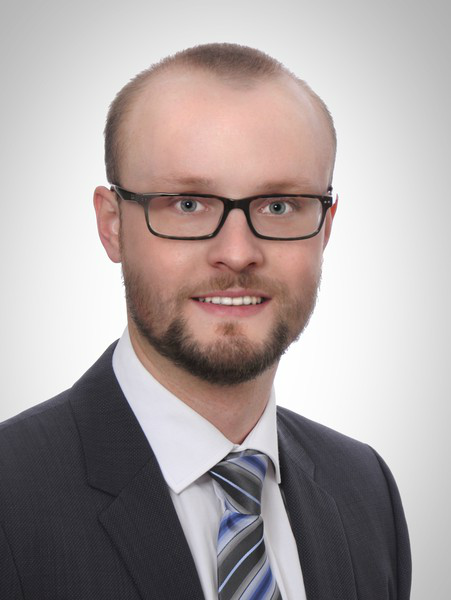}}]{Tobias Ross}
received his MSc in Medical Informatics from the University of Heidelberg/Heilbronn. His masters thesis was about crowd-algorithm collaboration for large-scale image-annotation. He is currently working in the Division of Computer Assisted Medical Interventions at the German Cancer Research Center (DKFZ) in Heidelberg as a Phd student. His research focuses on robotics and surgical workflow analysis.
\end{IEEEbiography}

\begin{IEEEbiography}[{\includegraphics[width=1in,height=1.25in,clip,keepaspectratio]{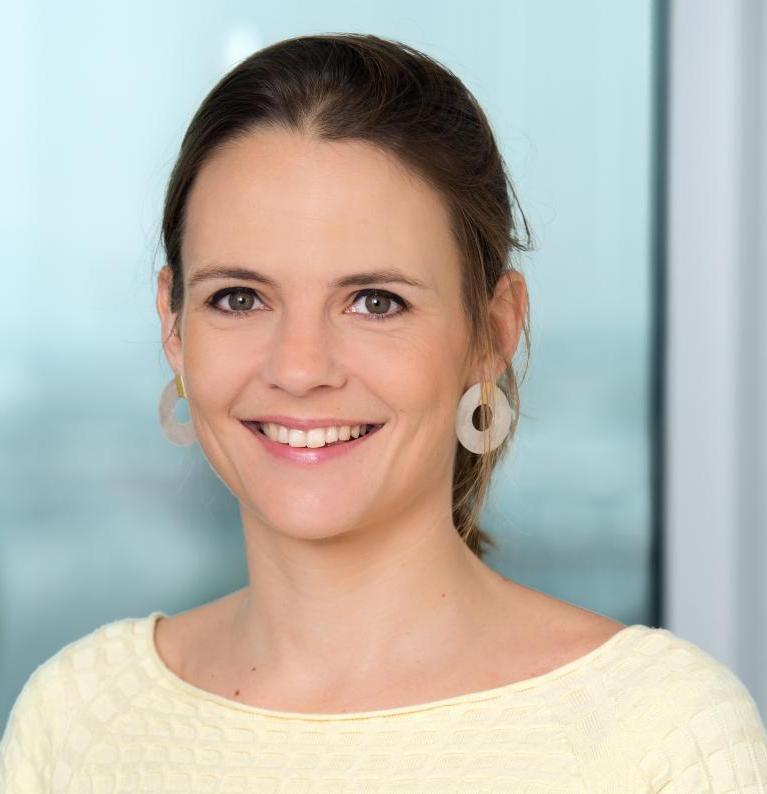}}]{Lena Maier-Hein}
received her PhD from Karlsruhe 
Institute of Technology KIT with distinction in 2009 and conducted her postdoctoral 
research in the Division of Medical and Biological Informatics at the German Cancer 
Research Center (DKFZ) and at the Hamlyn Centre for Robotics Surgery at Imperial 
College London. As a full professor at the DKFZ, she is now working 
in the field of computer assisted medical interventions, focusing on multi-modal image processing, surgical data science and computational biophotonics. She has and 
continues to fulfill the role (co-) principal investigator on a number of national and 
international grants including a European Research Council (ERC) starting grant 2014.
\end{IEEEbiography}
\vfill




\end{document}